\documentclass[aps,preprint,showpacs,superscriptaddress]{revtex4-1}  
\usepackage{graphicx}
\usepackage{subcaption}
\usepackage{bm}   
\usepackage{color}
\usepackage{amssymb}
\usepackage{amsmath}
\usepackage{float}

\begin{document}
	
	\title{Model-free prediction of emergence of extreme events in a parametrically driven nonlinear dynamical system by Deep Learning}
	\author{J.Meiyazhagan}
	\affiliation{Department of Nonlinear Dynamics, Bharathidasan University, Tiruchirappalli - 620 024, Tamilnadu, India}
	\author{S. Sudharsan}
	\affiliation{Department of Nonlinear Dynamics, Bharathidasan University, Tiruchirappalli - 620 024, Tamilnadu, India}
	\author{M. Senthilvelan}
	\email[Correspondence to: ]{velan@cnld.bdu.ac.in}
	\affiliation{Department of Nonlinear Dynamics, Bharathidasan University, Tiruchirappalli - 620 024, Tamilnadu, India}
	\vspace{10pt}
	
	\begin{abstract}
		We predict the emergence of extreme events in a parametrically driven nonlinear dynamical system using three Deep Learning models, namely Multi-Layer Perceptron, Convolutional Neural Network and Long Short-Term Memory. The Deep Learning models are trained using the training set and are allowed to predict the test set data. After prediction, the time series of the actual and the predicted values are plotted one over the other in order to visualize the performance of the models. Upon evaluating the Root Mean Square Error value between predicted and the actual values of all three models, we find that the Long Short-Term Memory model can serve as the best model to forecast the chaotic time series and to predict the emergence of extreme events for the considered system.
	\end{abstract}
	
	%
	%
	%
	%
	%
	\maketitle
\section{Introduction}
\par Extreme events are rare and recurrent events that occur in nature. They do appear in the form of rogue waves, earthquakes, floods, tornadoes and as epidemics. \cite{dysthe20081,jentsch2005,krause20151}. It is evident that each one of them can produce highly destructive consequences when they appear. During the past two decades considerable efforts have been made to understand certain basic characteristics about these events both theoretically and through experiments performed in laboratories. In the study of extreme events, four important points are considered, namely (i) prediction, (ii) mechanism, (iii) mitigation and (iv) statistics {\cite{fara2008}}. Among these, the prediction of extreme events play a vital role since it helps to determine the other three. Prediction is the method of determining the value of some observables at a specific time in future from the available data of the past. Even though classical prediction is an easily achievable task in simple systems, it turns out to be a difficult one in the case of chaotic and complex systems. To overcome this difficulty, in recent years, Machine Learning (ML) algorithms have been predominantly used for the prediction tasks. The main two prediction tasks used in ML are (i) classification and (ii) regression. In classification, the algorithm attempts to classify the input data among the preset categories whereas in the regression type, the algorithm predicts the output values by learning continuous mapping functions from the various input features \cite{goodfellow2016}.
\par These two prediction tasks are  utilized in almost all branches of physics in order to make model-free prediction of the future values of certain observables \cite{Lohani2020,Carleo2019,Radovic2018}. In recent years, ML algorithms are used as a viable tool to probe various characteristics of dynamical systems. For example, using ML algorithms, chaotic attractors are replicated in \cite{Pathak2017}, unstable periodic orbits are detected in \cite{zhu2019}, chaotic signals are separated in \cite{Krishnagopal2020}, the amplitude of chaotic laser pulses are predicted in \cite{Amil2019} and chimera states are identified in \cite{BARMPARIS2020,ganaie2020,kushwaha2020}. In addition to this, in the analysis of prediction of extreme events using ML models, reservoir computer (RC) has been used in a coupled FitzHugh-Nagumo model \cite{PYRAGAS2020}, Artificial Neural Network (ANN) has been used in H\'enon map \cite{lellep2020}, \textit{k}-means clustering and ANN have been used in optical fibre modulation instability \cite{Narhi2018} and wavelet-based ML methods have been used to forecast the extreme flood events in the flood-prone river basin \cite{Yeditha2020}.


\par In all the above works, ML methods have been used for the model-free prediction of the observables of the underlying physical system. To obtain a better model-free prediction, we can utilize the Deep Learning (DL) methods which are subset of ML. Three well-known DL models in this category are (i) Multi-Layer Perceptron (MLP), (ii) Convolutional Neural Network (CNN) and (iii) Long Short-Term Memory (LSTM). Among these MLP model is a normal ANN having more than one hidden layer, CNN models are working on the principle of convolution operation and LSTM models are neural networks that are designed to remember the information of the given data for a longer period of time. As prediction is an important task in the study of extreme events, in the present work, we utilize these three DL models to make a model-free prediction of the emergence of extreme events by forecasting the time series of a non-polynomial dynamical system with velocity dependent potential which is known for its rich nonlinear dynamics \cite{ss1}. We mention here that differing from the conventional applications, the DL models mentioned above have also been used in forecasting the future values of time series data in certain fields \cite{SEZER2020,gamboa2017,SHEN2020}. Other than DL models, RC has been used to predict the time series of chaotic food chain system, power system model and 1D Kuramoto-Sivashinsky system \cite{kong2021machine,wikner2020combining}, while, transfer learning was used to forecast the time series of Lorenz map \cite{ye2021implementing}. Further RC was also used to predict time series of Mackey–Glass differential delay equation in \cite{bollt2021explaining,canaday2018rapid}.

In the present work, we predict the occurrence of extreme events with the aid of MLP, CNN and LSTM. We show that LSTM can predict the emergence of extreme events with high accuracy. To the best of our knowledge this is the first time these three DL models are used for the model-free prediction of the emergence of extreme events in a highly nonlinear and chaotic system.

\par We organise the paper in the following way. In Sec. 2, we introduce the nonlinear system and data preparation for the prediction of extreme events using DL models. In Sec. 3, we present the working principles of three DL models. In Sec.~4, we discuss the behaviour of each model based on its performance and analyze the results of the DL models. We also test the memory of LSTM using parameter switching in this section. Finally, we present our conclusion in Sec. 5.

\section{System and Data preparation}
\begin{figure}[!ht]
	\centering
	\includegraphics[width=0.5\linewidth]{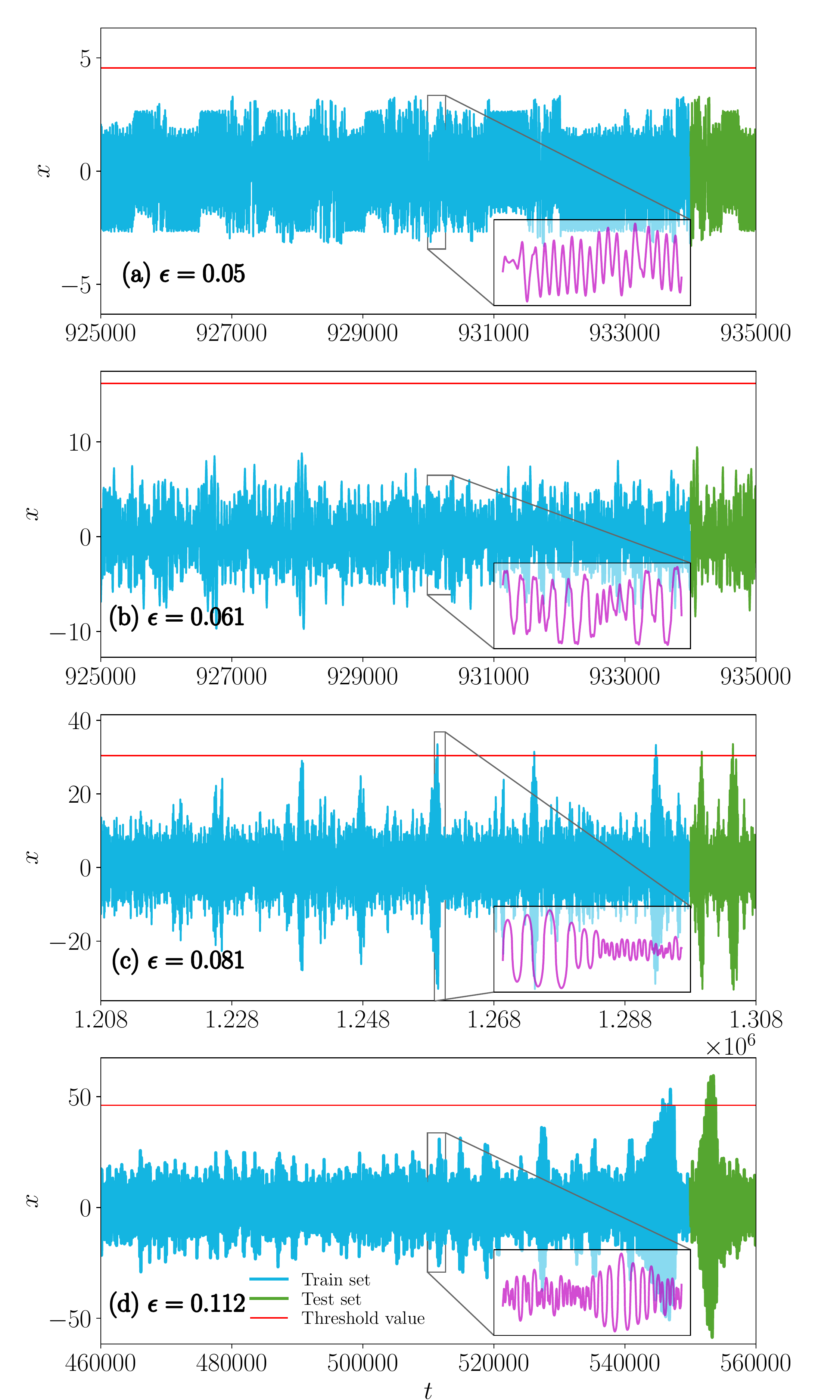}
	\caption{\label{fig1} Time series values of the four different values of $\epsilon$. (a) and (b) are non-extreme event cases and (c) and (d) are extreme event cases. The inset in the subplots represents the trajectories for a very small time domain which confirms the chaotic nature of the trajectory.}
\end{figure}

We consider a mechanical model which describes the motion of a freely sliding particle of unit mass on a parabolic wire rotating along the axis of rotation $z = \sqrt{\lambda} x^2$ with a parametrically varying angular velocity $\Omega=\Omega_0(1+\epsilon \cos \omega_pt)$ where $\lambda > 0$. Here $\epsilon$ and $\omega_p$ are the strength and frequency of the external drive. The corresponding equation of motion, with linear damping is given by \cite{ss1}

\begin{equation}
(1 + \lambda x^2) \ddot{x} + \lambda x \dot{x}^2 + \omega_0^2 x - \Omega_0^2 \left[ 2\epsilon \cos \omega_pt + \frac{1}{2} \epsilon^2 (1 + \cos 2\omega_pt) \right] x + \alpha \dot{x} = 0, 
		\label{pardri}
\end{equation}
where $\frac{1}{\sqrt{\lambda}}, \omega_0$ and $\alpha$ are semi-latus rectum of the rotating parabola, initial angular velocity and the  positive  linear  damping respectively. The parameters are fixed as $\lambda=0.5, \omega_0^2 = 0.25, \Omega_0^2=6.7, \omega_p = 1.0, \alpha=0.2$ and $\epsilon$ is the bifurcation parameter. The emergence of extreme events in the system (\ref{pardri}) has been dealt very recently in \cite{ss1}. In this system, an event is said to be extreme if the trajectory of $x$ or $\dot{x}$ seldom crosses a calculated threshold value $x_{ee}$. This threshold value is calculated using the formula $x_{ee}=\langle x\rangle+4\sigma_x$, where $\langle x\rangle$ is the mean peak amplitude and $\sigma_x$ is the standard deviation. Here we intend to predict the time series of (\ref{pardri}) using DL models both in the non-extreme event regimes and in the extreme event regimes. We plot the time series of the above mentioned system in Fig. \ref{fig1} for the values $\epsilon=0.05,0.061,0.081$ and $0.112$. The red horizontal line in Fig. \ref{fig1} represents the calculated threshold value. Here $\epsilon=0.05~\text{and}~0.061$ (Figs. \ref{fig1}(a) \& (b)) represent the parameters corresponding to the non-extreme event regimes, whereas $\epsilon=0.081~\text{and}~0.112$ (Figs. \ref{fig1}(c) \& (d)) correspond to the extreme event regimes. The data in the time series are split up into two sets, namely training set and test set which contains $18000$ and $2000$ data respectively. The training and test sets are respectively represented by blue and green trajectories in Fig.~{\ref{fig1}}. The inset in the subplots (a)-(d) in {Fig.~\ref{fig1}} represent the trajectories of a very small time domain and they are shown to confirm the chaotic nature of the trajectory.
\par In Figs.~\ref{fig1}(a) \& (b) we can see that there is no emergence of extreme events in the training and test set data for $\epsilon=0.05,0.061$. But in Figs. \ref{fig1}(c) \& (d) extreme events are present both in the training and test set data. With this training set data, we examine the ability of DL models to learn the nonlinear behaviour of the system \eqref{pardri} and allow the models to predict the future values of the time series. Besides forecasting the time series, the model also predicts the emergence of extreme events which can be inferred from the predicted time series values. The performance of each DL model is determined by comparing the predicted values with test set values. Before applying DL models on the data-sets, we convert the time series data into a supervised learning data. In this connection, we take $x(t)$ at each step as a series,
\begin{equation}
	 x_1, x_2, x_3, ... x_n.
\end{equation} 
This can be converted to supervised learning data by taking $x_n$ as input and $x_{n+1}$ as the corresponding output. The notation $x_1, x_2, x_3, ... x_n$ denotes the values of $x$ at $t=1,2,3,...n.$

\begin{table}[!ht]
	\centering
	\begin{tabular}{|c|c|}
		\hline
		Input $(X)$ & Output $(Y)$ \\
		\hline
		$x_1$ & $x_2$ \\
		\hline
		$x_2$ & $x_3$\\
		\hline
		$x_3$ & $x_4$\\
		\hline
		\vdots & \vdots \\
		\hline
		$x_{n-1}$ & $x_n$\\
		\hline
	\end{tabular}
	\caption{\label{table1}Demonstration of changing the time series data into supervised learning data }
\end{table}
\par Both the training and test set are rescaled using \textit{min-max} normalization. The formula for \textit{min-max} normalization between the range $\alpha$ and $\beta$ is given by \cite{al2006},
\begin{equation}
	x_i^{rescaled} = \alpha + \dfrac{(x_i-x_{min})(\beta-\alpha)}{x_{max}-x_{min}},\quad i=1,2,3,..,n,
\end{equation}
where $x_{min}$ and $x_{max}$ are respectively the minimum and maximum value of the data set $X$. Conventionally, we fix $\alpha=-1$ and $\beta=1$ to scale the values between -1 and +1. In order to compare the predicted values with the actual values, we will reverse all the preprocessing steps after obtaining the output from the DL model. 
\section{Deep learning models}
As mentioned earlier, we train the DL models, ANN, CNN and LSTM, to forecast the future values of the time series of the considered nonlinear system. To train these three DL models, we adopt the following procedure. To begin, we feed the input values  $X_{train}$ and $Y_{train}$ into the above three DL models. In each DL model, we perform the training process for 250 epochs (here epoch indicates the number of times that the entire training data set has been passed into the learning algorithm). Next we forecast future values of the time series of the considered system with trained DL models. We use the walk forward validation method \cite{walk} to predict the future steps of the time series (test set data). In this method, the predicted value from the output is added into the input at each step of forecasting.

In the following, we describe the considered models one by one and point out how each one of them predicts the emergence of extreme events in the system (\ref{pardri}).  

\subsection{Multi-Layer Perceptron}
\begin{figure}[!ht]
	\centering
	\includegraphics[width=0.5\linewidth]{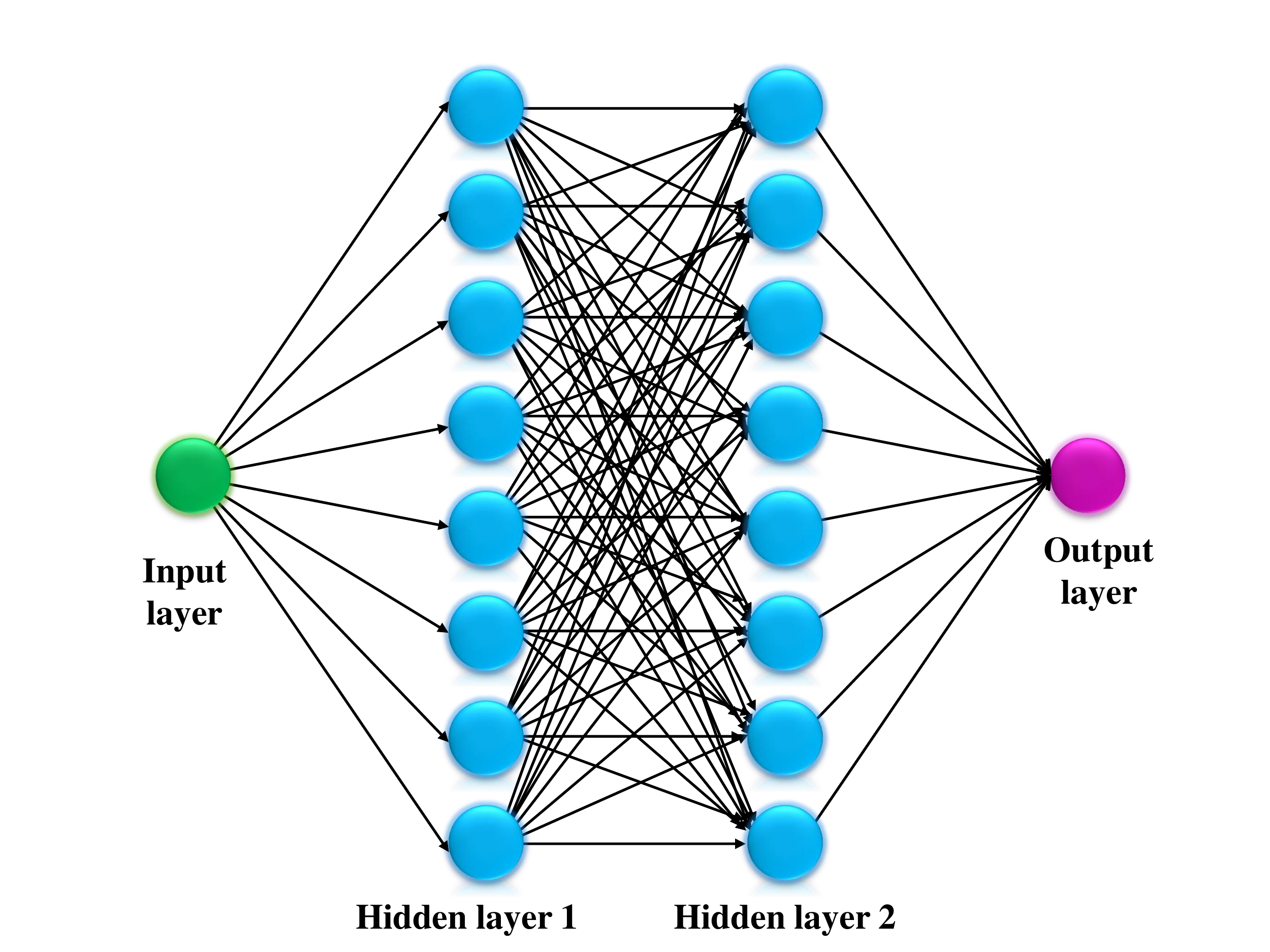}
	\caption{\label{mlp} Schematic diagram of MLP}
\end{figure}

An ANN which has more than one hidden layer is often called as MLP \cite{goodfellow2016}. The schematic diagram of MLP is shown in Fig.~\ref{mlp}. The input $X$ is fed through the input layer while the output $\hat{Y}$ is obtained from the output layer. In between there are two hidden layers each consisting of eight neurons. All the neurons in the MLP model are fully connected with each other.
\par
\begin{figure}[!ht]
	\centering
	\includegraphics[scale=0.5]{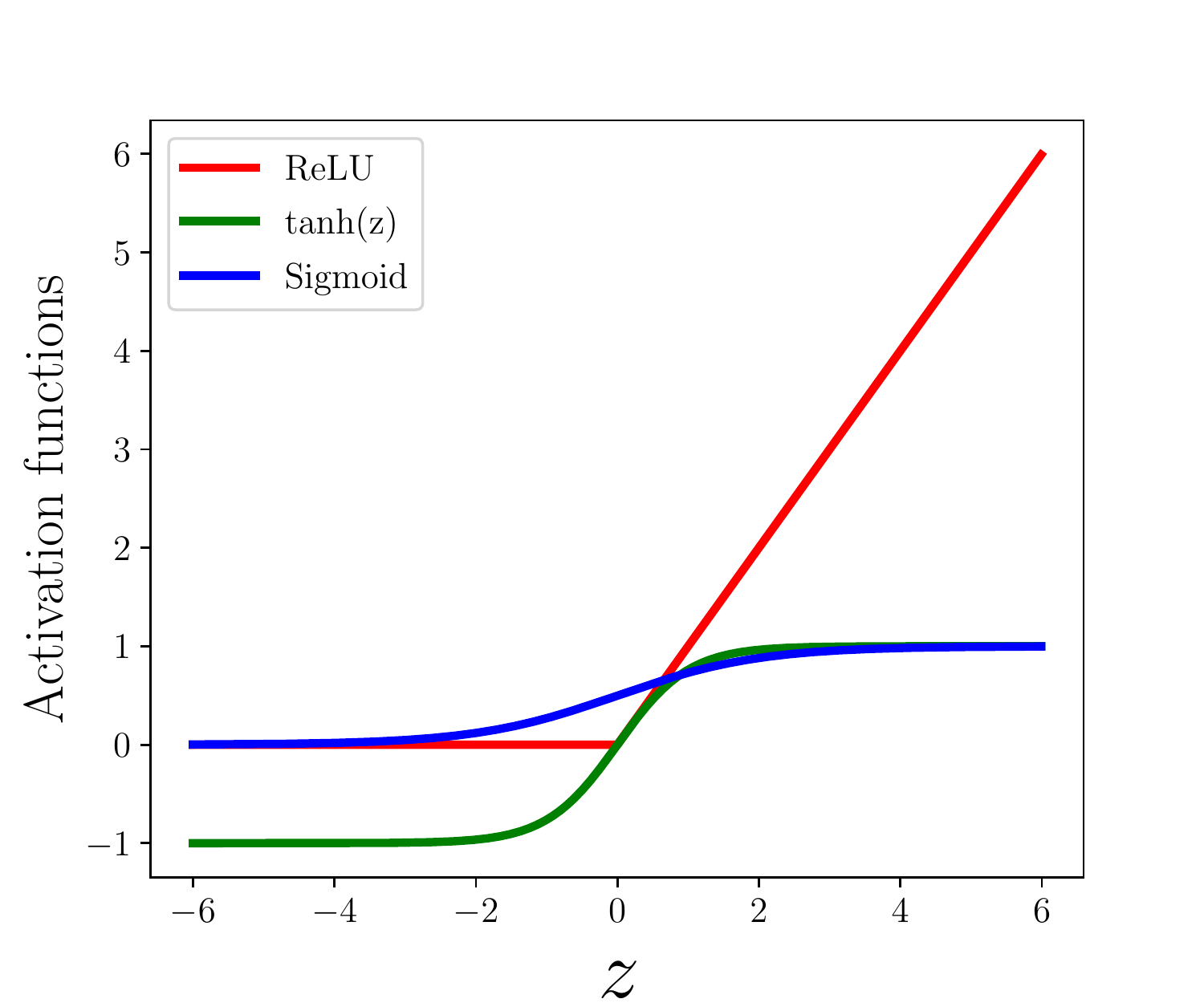}
	\caption{\label{Ac_fn} Plot of different types of activation functions. The red line represents the ReLU function for the activation of neurons. The green and blue line represents $\tanh$ and sigmoid functions respectively which are used in LSTM layers.}
\end{figure}

\begin{figure}[!ht]
	\centering
	\includegraphics[width=0.5\linewidth]{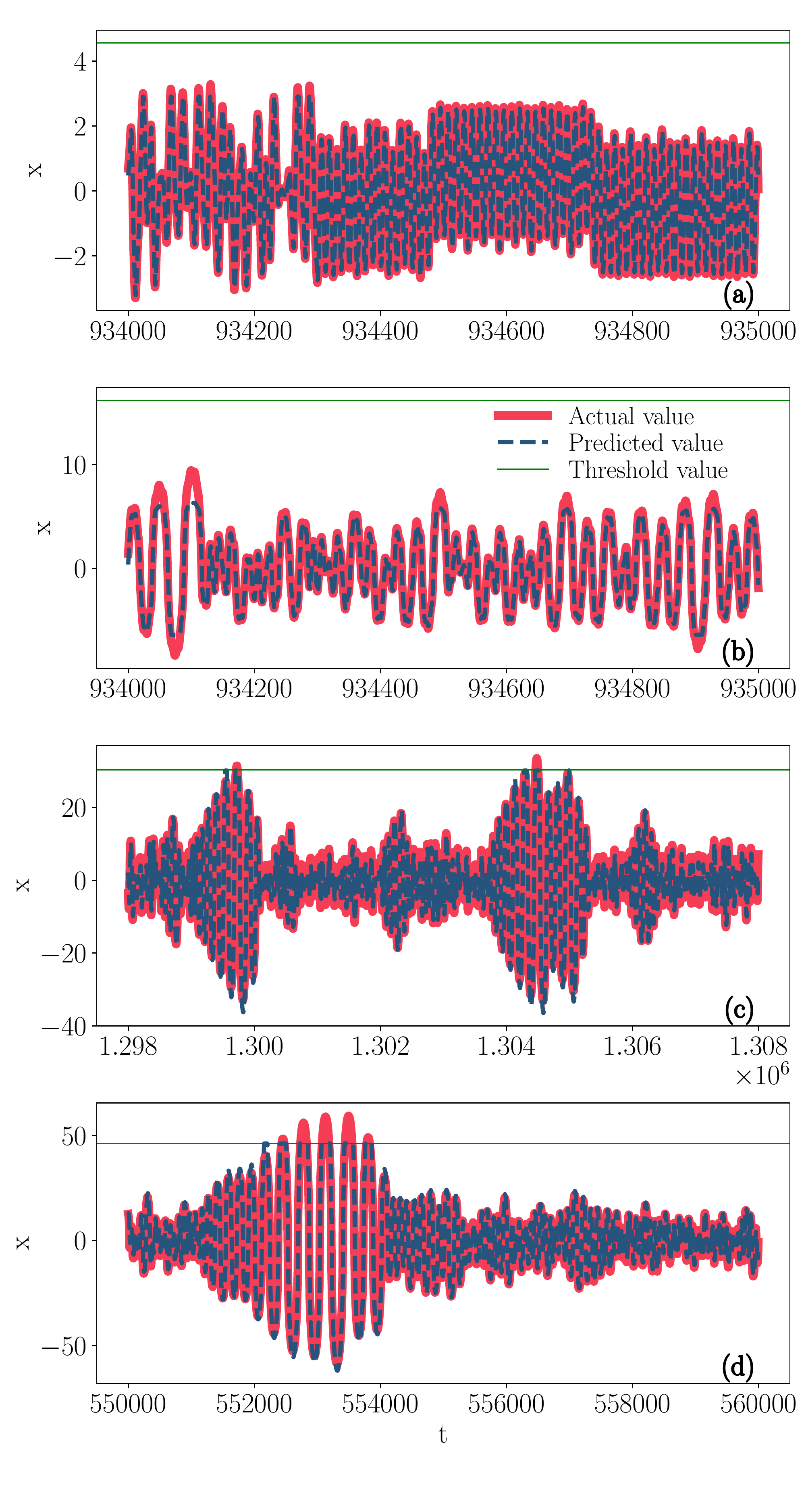}
	\caption{\label{mlp_result} Plots of predicted values of time series of each $\epsilon$ values over the actual values of the MLP model. The solid red line represents the actual test set data, dotted blue line represents the predicted value using MLP model and solid green line represents the calculated threshold value.}
\end{figure}

 The interactions between subsequent layers of MLP are determined by applying an activation function $a$ to the latent variable $z^{(l)}$, where $z^{(l)}$ takes the form 
\begin{subequations}
	\begin{equation}
	\textbf{z}^{(l)} = \textbf{W}^{(l,l-1)}.\textbf{y}^{(l-1)}+\textbf{b}^{(l)},
	\end{equation}
	\begin{equation}
	\textbf{y}^{(l)} = a(\textbf{z}^{(l)}),
	\end{equation}
\end{subequations}
where $\textbf{W}^{(l,l-1)}$ and $\textbf{b}^{(l)}$ are the weight and bias matrices respectively. The notations $(l)$ and $(l-1)$ at the superscripts indicate a layer and its previous layer respectively in the network. These matrices are arbitrarily changed during the training process in order to minimize the loss function $(L)$ of the network. In general, Mean Squared Error (MSE) is used as a loss function in regression problems in the form
\begin{equation}
	 L_{MSE} = \sum_{i=1}^{N_{Train}}\dfrac{(\hat{Y}^{Train}_i-Y_i^{Train})^2}{N_{Train}},
\end{equation}
where $\hat{Y}^{Train}_i,Y_i^{Train}$ and $N_{Train}$ are the predicted values, actual values and total number of data in the training set respectively. Further, Adam optimizer \cite{kingma2014} is used to minimize the $L_{MSE}$. For all the hidden layers in the network, the  Rectified linear activation unit (ReLU) \cite{goodfellow2016} is used as an activation function $(a(\textbf{z}))$ in the form
\begin{equation}
 a_{ReLU}(z) = max(0,z).
\end{equation} 
The working behaviour of this activation function can be easily visualized from Fig.~\ref{Ac_fn}. 
Now the training set is fed into the MLP model for the learning process. In our case after running 250 epochs the model is allowed to predict the values of the test set data. To see the performance of the MLP model, we plot the predicted data over the actual test set data and present the outcome in Fig. \ref{mlp_result}.
The solid red line in Fig. \ref{mlp_result} represents the actual test set data, dotted blue line represents the predicted value using MLP model and solid green line represents the calculated threshold value. Figures \ref{mlp_result}(a) and (b) are the time series plots corresponding to the values of $\epsilon=0.05$ and $0.061$ representing the non-extreme event regimes, while Figs. \ref{mlp_result}(c) and (d) are the time series plots corresponding to the values of $\epsilon=0.081$ and $0.112$ representing the extreme event regimes. The prediction made using the created MLP model is found to be in good agreement with the actual test set data in both non-extreme event and extreme event cases.
\subsection{Convolutional Neural Network}
\begin{figure*}[!ht]
	\centering
	\includegraphics[width=0.8\linewidth]{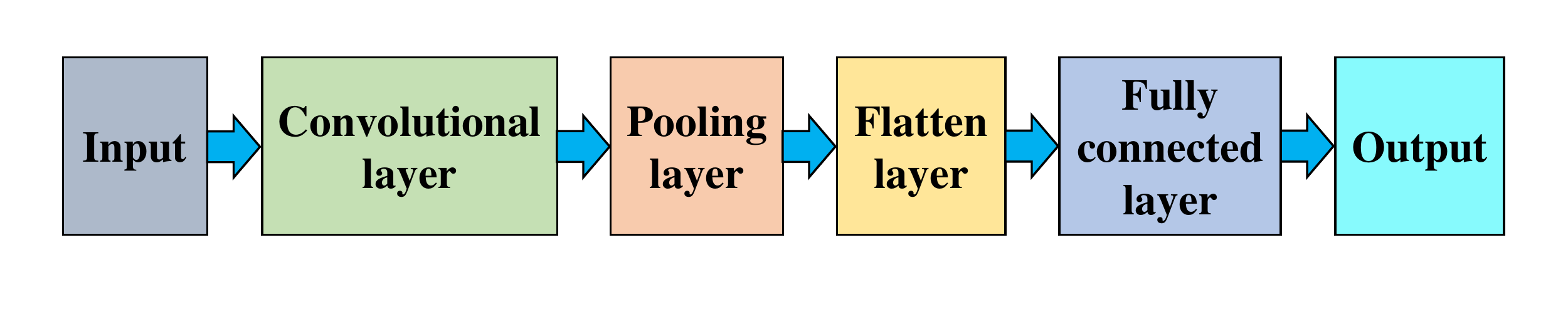}
	\caption{\label{cnn} Working pipeline diagram of CNN}
\end{figure*}

CNN is another type of neural network often used in image processing and time series forecasting \cite{borovykh2018}. The term `\textit{convolutional}' indicates that this model works on the principle of mathematical convolution operation. The CNN models are most commonly used in the field of computer vision such as image classification, object detection and image style transfer \cite{Gatys_2016_CVPR}. Pixel values of the image are taken as $n_H\times n_W \times n_C $ matrix, where $n_H$ and $n_W$ are number of pixels at height and width of the image and $n_C$ is the total number of channels of image. For gray-scale image  $n_C=1$ and for RGB image $n_C=3$ \cite{goodfellow2016}. The matrix which is used for convolution operation in the input matrix is known as kernel. One dimensional CNN model has been used to forecast the time series in the areas such as Electricity Load Forecasting \cite{1dcnn2019} and financial time series \cite{cnn2020}. Based on this observation, in our studies, we train this DL model to forecast the future values of the time series of the system \eqref{pardri}. The working pipeline diagram of CNN is shown in Fig. \ref{cnn}. 
\par In the CNN, the time series is first fed into the convolutional layer where the convolution operation is performed by applying 64 filters of kernel size one. Followed by this a max pooling layer is added which takes the maximum value of the given data for the applied filter size. Next a fully connected layer with 50 neurons is used. A flatten layer is used in between the pooling and fully connected layer. Finally, an output layer with one neuron is connected. In this model also we use ReLU activation function for all layers and Adam optimizer to minimize the $L_{MSE}$.
\par
\begin{figure}[!ht]
	\centering
	\includegraphics[width=0.5\linewidth]{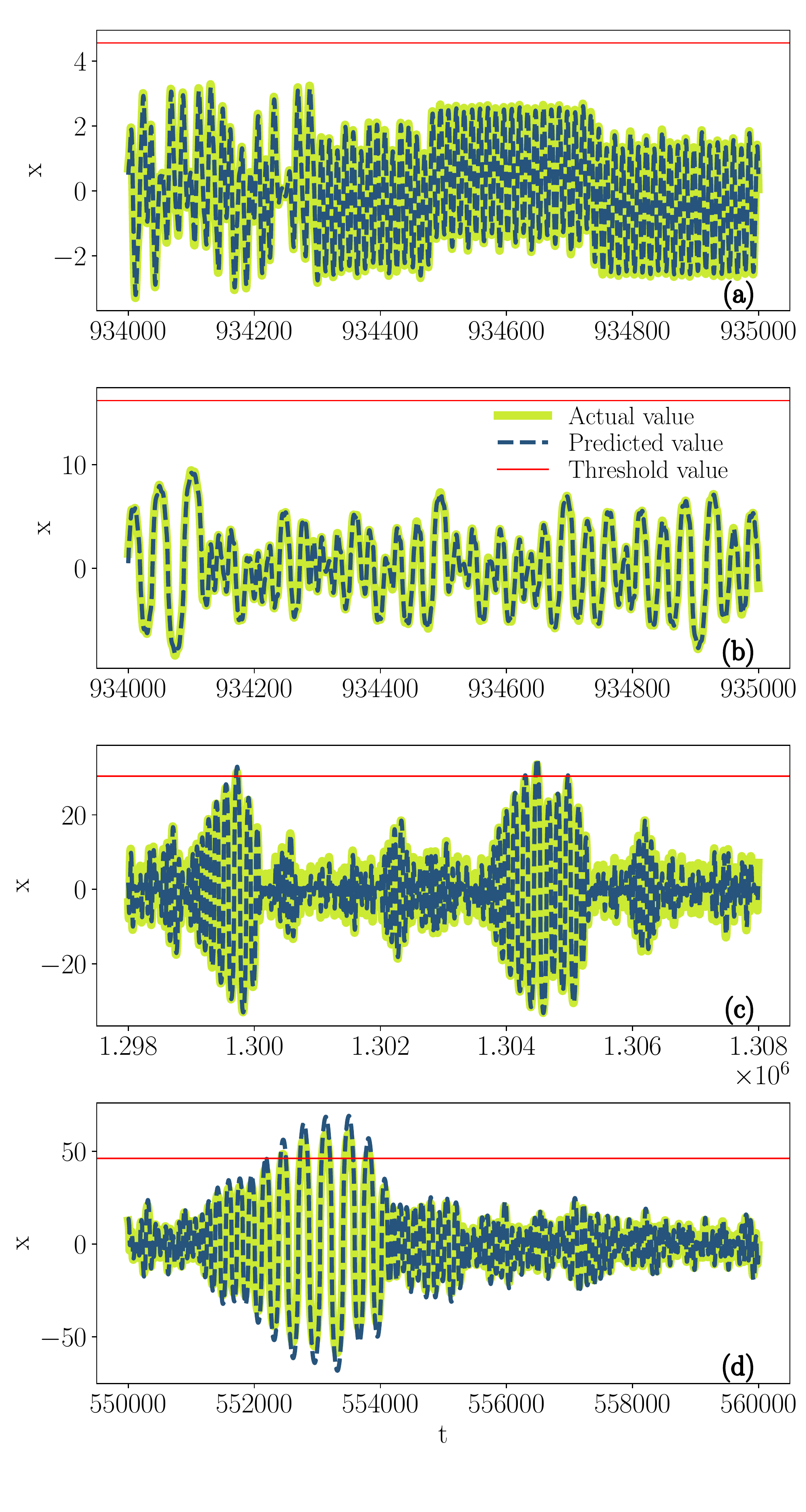}
	\caption{\label{cnn_result}Plots of predicted values of time series over the actual values of the CNN model for four different values of $\epsilon$. The solid green line represents the actual test set data, dotted blue line represents the predicted value using CNN model and solid red line represents the calculated threshold value.}
\end{figure}

 After training, this CNN model is allowed to predict the test set data as in the previous case.  Actual and predicted values are plotted in Fig. \ref{cnn_result}. 
In this figure, solid green line represents the actual test set data, dotted blue line represents the predicted value using CNN model and solid red line represents the calculated threshold value. In CNN, we find a good agreement between the actual and predicted values.
\subsection{Long Short-Term Memory}
Recurrent Neural Networks (RNN) \cite{rumelhart1986} are mostly employed when the input has a sequence of data. This type of networks have loops in them in order to allow the information to persist. It is used in speech recognition, sentiment classification, video activity recognition, music creation \cite{hadjeres2020} and in the analysis of DNA sequence \cite{liu2019}. LSTM networks are a special kind of RNNs \cite{lstm1}. It is used in forecasting the time series of systems that have high dependency on the past values \cite{chimmula2020}. 
Remembering information for longer periods of time is the default behaviour of LSTM networks. All RNN models have the chain of repeating modules of the neural networks. The standard RNN models use only one hyperbolic tangent $(\tanh)$ activation function. But in the LSTM, $\tanh$ is used for the activation and sigmoid function is used for recurrent activation. The functionality of these two functions can be easily understood from the Fig. \ref{Ac_fn}. The general form of the sigmoid function is given by, 
\begin{equation}
	\qquad\qquad\qquad\sigma(z) = \dfrac{1}{1+e^{-z}}.
\end{equation}

\par
\begin{figure}[!ht]
	\centering
	\includegraphics[width=0.5\linewidth]{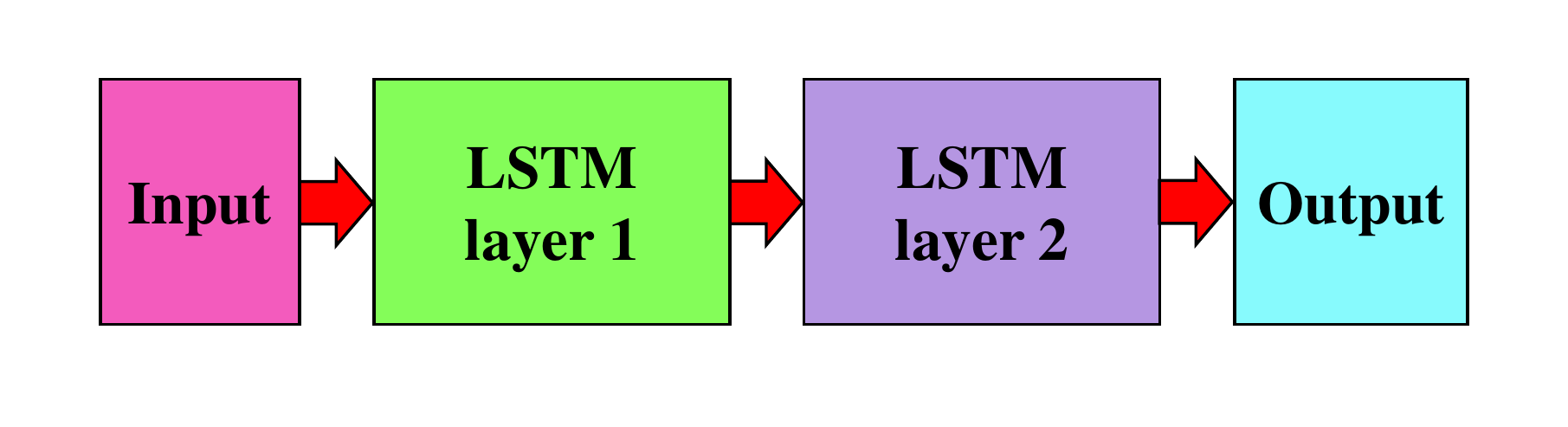}
	\caption{\label{lstm} Working pipeline diagram of LSTM}
\end{figure}

\begin{figure}[!ht]
	\centering
	\includegraphics[width=0.5\linewidth]{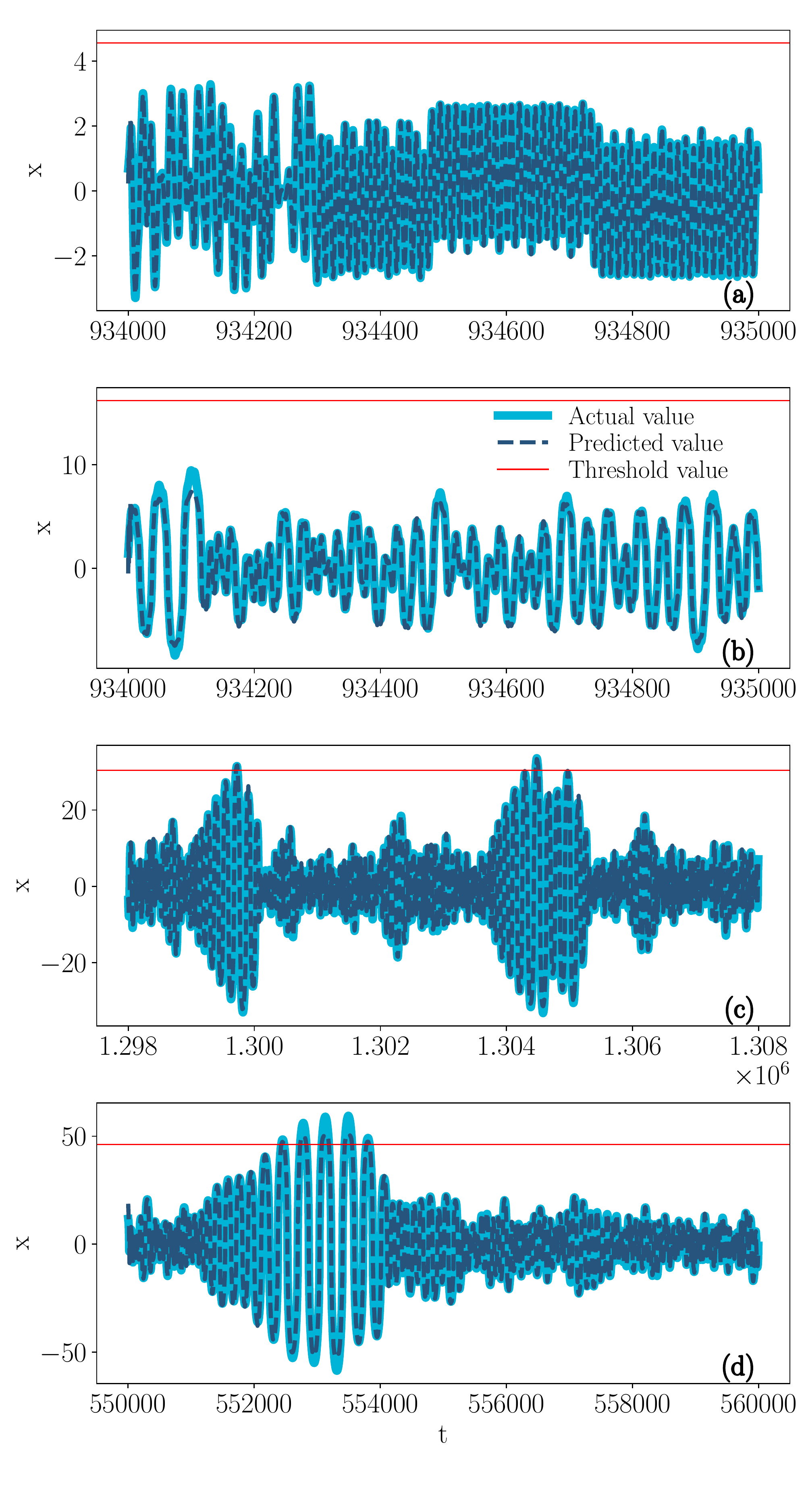}
	\caption{\label{lstm_result} Plots of predicted values of time series of each $\epsilon$ values over the actual values of the LSTM model. The solid light blue line represents the actual test set data, dotted dark blue line represents the predicted value using LSTM model and solid red line represents the calculated threshold value.}
\end{figure}

 The working pipeline diagram of our LSTM model is shown in the Fig. \ref{lstm}. 
The input is first given to the LSTM layer 1 and subsequently to LSTM layer 2 in order to find the long term dependencies between the data. Both LSTM layers have 32 units. An output layer with a neuron is connected to the second LSTM layer. This model is trained using the training set data for all the four cases. As mentioned earlier we run this model for 250 epochs. To evaluate the performance of this model, the test set data is used to forecast the time series for the considered problem. The predicted and actual values are plotted in Fig. \ref{lstm_result}.

In Fig. \ref{lstm_result}, solid light blue line represents the actual test set data, dotted dark blue line represents the predicted value using LSTM model and solid red line represents the calculated threshold value. In this case also, we find a good agreement between the actual and predicted values.

\subsubsection{Forecasting using the parameter values}
\begin{figure}[!ht]
	\centering
	\includegraphics[width=0.5\linewidth]{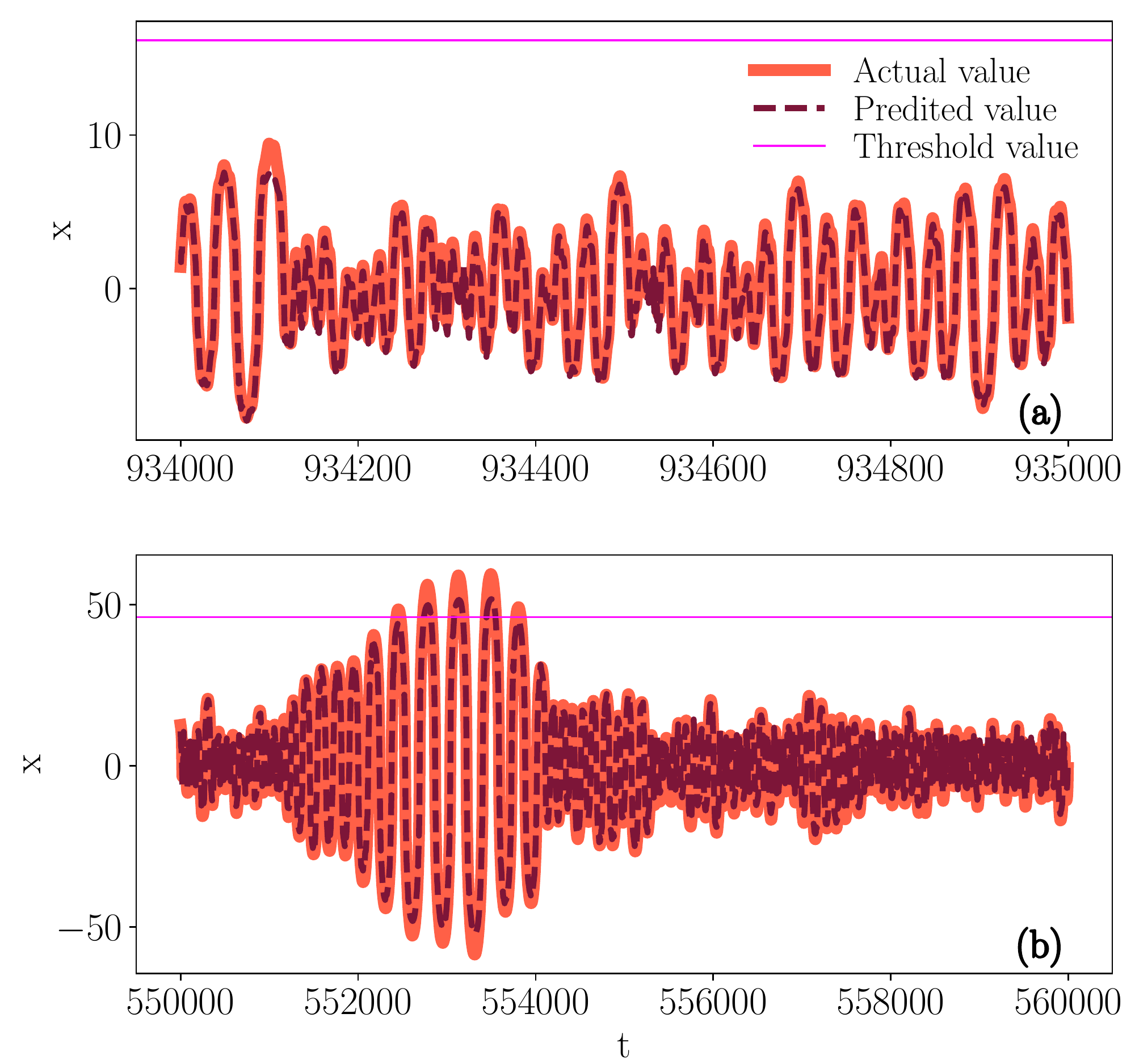}
	\caption{\label{mem_lstm_result} Plots of predicted values of time series over the actual values of the LSTM model trained using parameter values. The solid light orange line represents the actual test set data, dotted dark brown line represents the predicted value using LSTM model and solid violet line represents the calculated threshold value. Fig. (a) corresponding to the output for $\epsilon=0.061$ trained using the parameter value of $\epsilon=0.05$. Fig. (b) corresponding to the output for $\epsilon=0.112$ trained using the parameter value of $\epsilon=0.081$.}
\end{figure} 
Now we test the memory of LSTM by including the value of the parameter along with the input data during the training process itself. In this process, instead of testing the time series for the same parameter value, we switch the value of the parameter to a new value and then predict the time series corresponding to that new parameter value. By doing so, we can determine whether or not the model LSTM has learned the inherent nonlinear behaviour of the considered system. While implementing this, first we take the non-extreme regime time series ($\epsilon=0.05$) and insert the value of $\epsilon$ at each step in the input during the training process. After training, we forecast the time series of another parameter (say $\epsilon=0.061$). Similarly for extreme regime case first we take $\epsilon=0.081$ for training and then forecast the time series of $\epsilon=0.112$. The outcome of this experiment is shown in Fig.~{\ref{mem_lstm_result}}. From the obtained results we confirm that the LSTM model has the ability of storing the nonlinear behaviour of the considered system for the given parameter values in both extreme and non-extreme cases.

\section{Discussion}
\begin{figure}[!ht]
	\centering
	\includegraphics[width=0.5\linewidth]{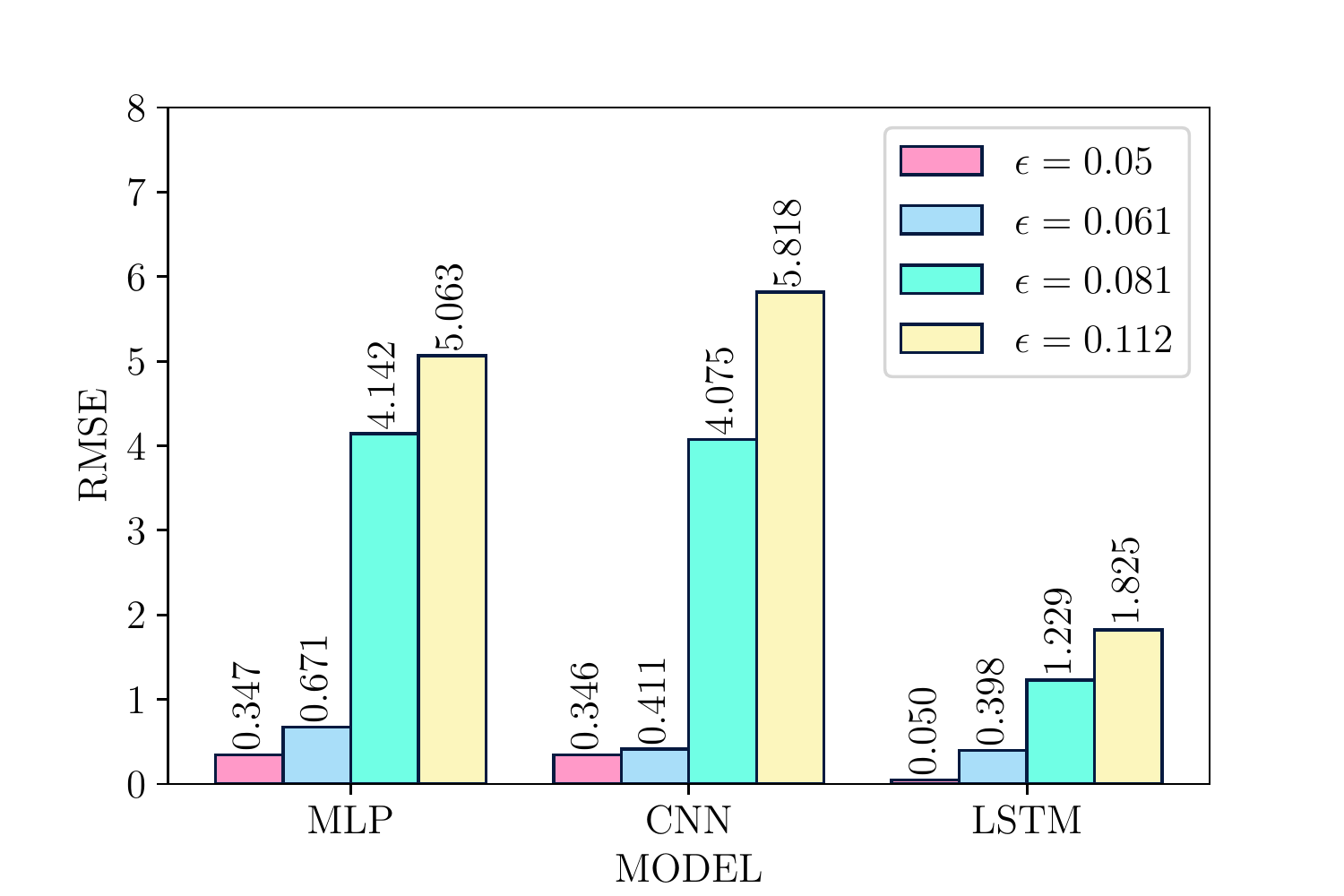}
	\caption{\label{bar} Bar plot of RMSE values for each DL model for non-extreme $(\epsilon = 0.05, 0.061)$ and extreme cases $(\epsilon = 0.081, 0.112)$.}
\end{figure} 
\begin{figure*}[!ht]
	\centering
	\includegraphics[width=0.85\linewidth]{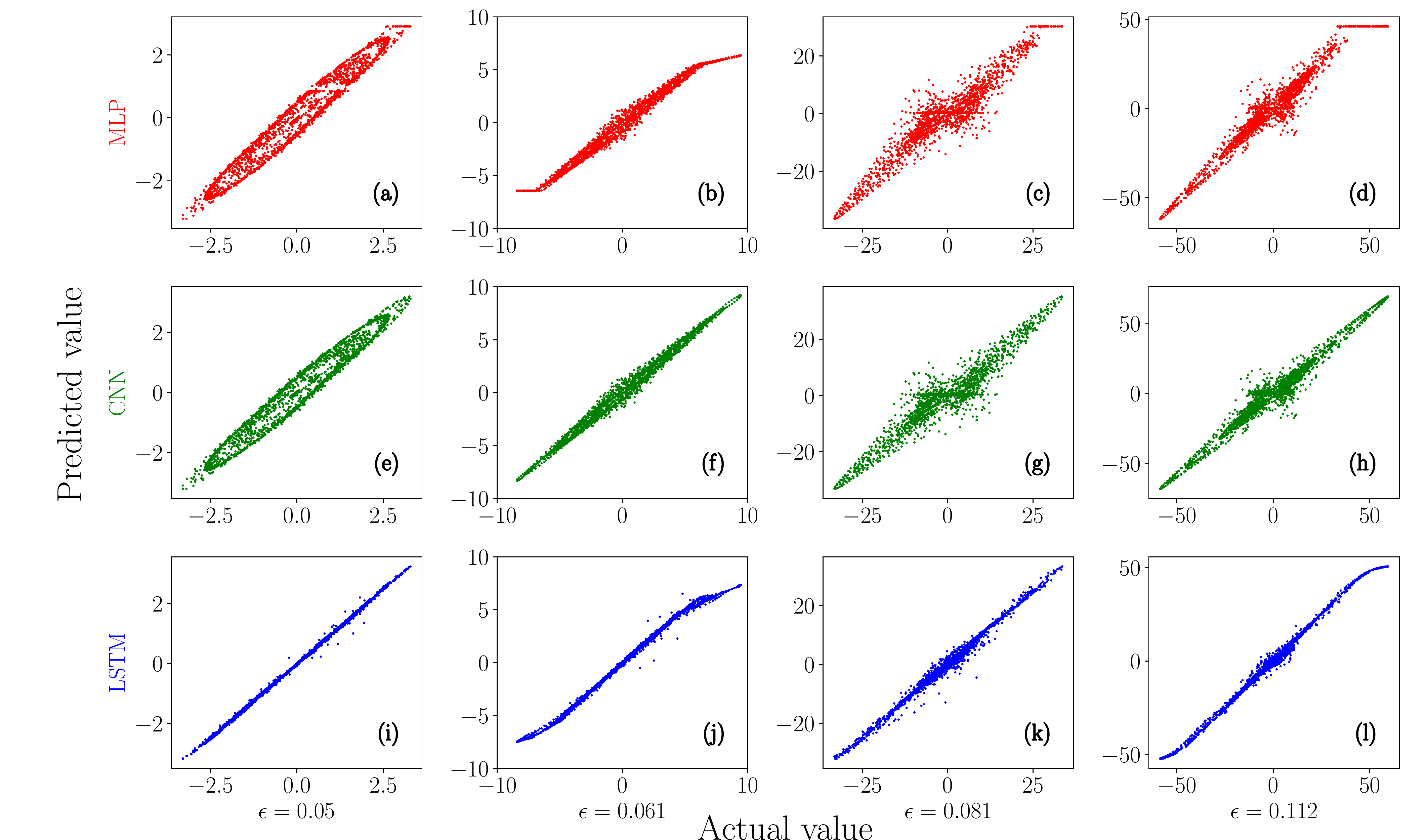}
	\caption{\label{scat} Scattered plot of actual values with predicted values of MLP, CNN and LSTM models}
\end{figure*}
After forecasting, to visualize the performance of the considered DL models, we evaluate the Root Mean Square Error (RMSE) value between predicted and actual values of the test set. The RMSE value is determined through the formula, 
\begin{equation}\label{rmse}
	\qquad\qquad\textnormal{RMSE} = \sqrt{\sum_{i=1}^{N_{Test}}\dfrac{(\hat{Y}^{Test}_i-Y_i^{Test})^2}{N_{Test}}},
\end{equation}
where $\hat{Y}^{Test}_i$, $Y_i^{Test}$ and $N_{Test}$ are the predicted values, actual values and total number of data in the test set respectively. The model which comes out with lower RMSE value is considered to be the suitable model for forecasting time series and predicting extreme events in the considered system.  Upon training the MLP model and evaluating its efficiency, we found that this model yields admissible RMSE values for the test set. Even though MLP model has offered a satisfactory prediction on the time series data, in order to reduce the RMSE value even further, we can opt other DL models, that is CNN and LSTM. The model with the least RMSE value predicts the time series of both extreme and non-extreme regimes more accurately. We observed that all three DL models have performed well in the prediction of emergence of extreme events and in fact the predicted values are considerably accurate. In order to determine the best DL model suitable for the prediction of extreme events in the system (\ref{pardri}), we have evaluated the RMSE value for each model and for all the four values of $\epsilon$. The RMSE value calculated from Eq. \eqref{rmse} are plotted as a bar graph in Fig. \ref{bar}.

\par From Fig. \ref{bar}, we can see that the RMSE value of MLP model (for all the four cases of $\epsilon$) is very high when compared with the other two models. While comparing LSTM with other two models, we can infer that it has the least RMSE value. We note here that in the literature, the RMSE value has been reported in the order of $10^{-2}$ for few algorithms, see for example Ref.~{\cite{ye2021implementing}}. On the other hand, in Ref.~{\cite{kong2021machine}}, the authors have obtained RMSE values in the range 0.07 to 0.17. In the present work, we have arrived at a comparable RMSE value for the problem under consideration. We also note that in the above works, forecasting was made to predict the time series in regular and chaotic systems without extreme events. But in our work, we forecast the time series of a highly nonlinear chaotic system with extreme values and found that the performance of all the three models is well and good. Upon comparison, we find that as far as the prediction of extreme events in system (\ref{pardri}) is concerned, we conclude that LSTM can serve as the best model among the considered three models.

\par To further confirm this, in Fig. \ref{scat}, we present the scatter plot of actual values (in $x$-axis) with the predicted values (in $y$-axis). 

The plot with less scattered points represents best fit and the plot with more scattered points represents the least fit. While comparing the scatter plots, we observe that scattering is high in MLP (Figs.~\ref{scat}(a)-(d)), moderate in CNN (Figs.~\ref{scat}(e)-(h)) and least in LSTM (Figs.~\ref{scat}(i)-(l)). This is due to the fact that LSTM model has the ability of storing information about the past values of the input time series data for longer periods of time. To validate this fact, we slightly modify the prediction task by additionally taking the value of the parameter corresponding to the time series in the input of the LSTM model. Then during the test, we forecast the time series corresponding to another parameter value. These results substantiate the fact that LSTM has a good ability in learning the nonlinear behaviour of the system. The obtained outcomes are examined through scatter plots, see Fig.~\ref{scat_mem}. We have included the complete details of the robustness of the models in Appendix. In particular, when we check the ability of DL models to make multi-step forecasting (see Appendix), we found that MLP and CNN models fail to make multi-step time series prediction whereas LSTM model enables us to make multi-step forecasting in both non-extreme and extreme cases. These confirmations also enforce us to conclude that LSTM is more accurate in predicting the emergence of extreme events in the considered system. 

\begin{figure}[!ht]
	\centering
	\includegraphics[width=0.5\linewidth]{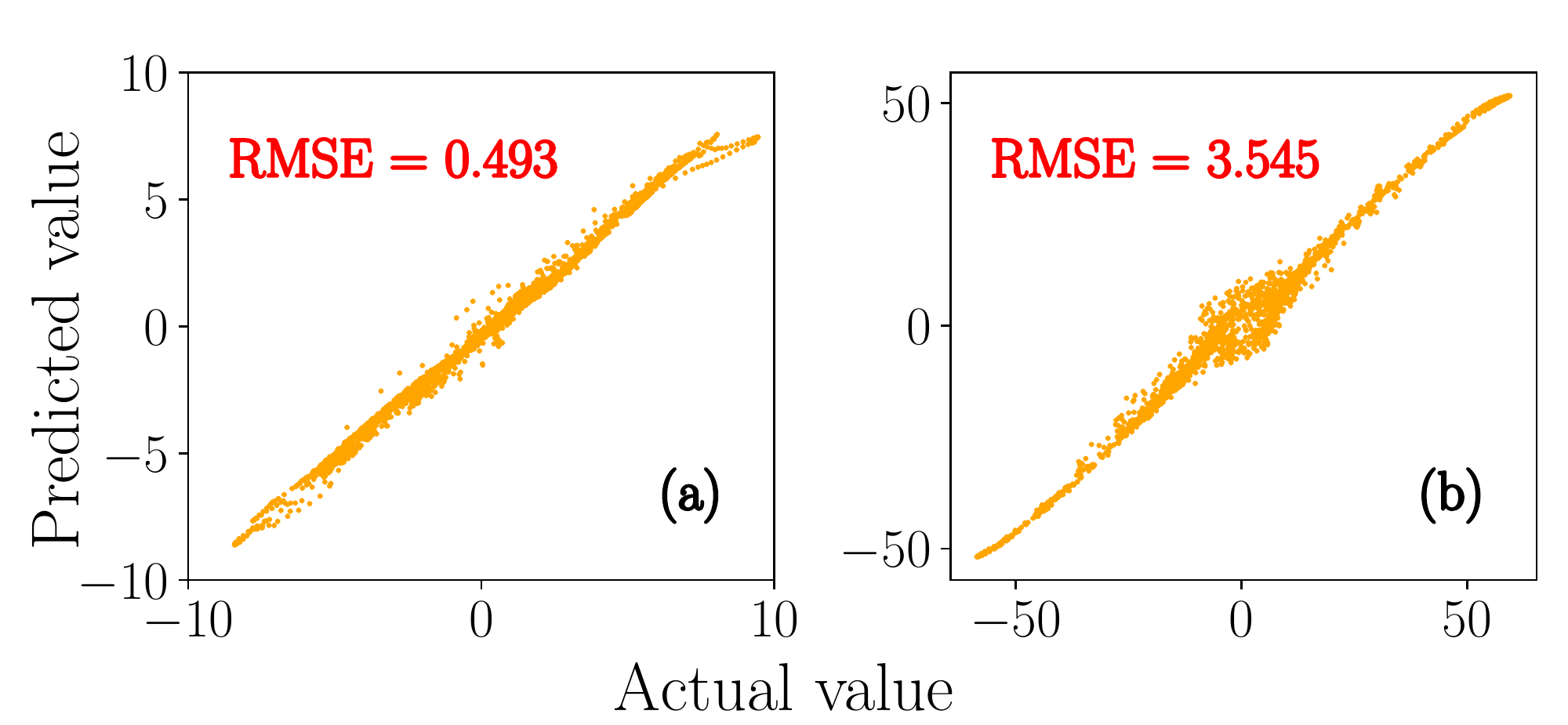}
	\caption{\label{scat_mem} Scattered plot of actual values with predicted values of LSTM trained using the parameter values. Fig.~(a) corresponds to the training using $\epsilon=0.05$ and testing for $\epsilon=0.061$ and Fig.~(b) corresponds to the training using $\epsilon=0.081$ and testing for $\epsilon=0.112$.}
\end{figure}
\section{Conclusion}
In this work, using DL models, we have predicted the occurrence of extreme events in a parametrically driven non-polynomial system with velocity dependent potential. In particular, we have considered the models MLP, CNN and LSTM to predict the extreme events. All the three models were made to learn from the training set of the time series data. Then the models were allowed to forecast the future values of the time series to predict the emergence of extreme event. This has been done for four different parameter values ($\epsilon$) of the considered dynamical system. Out of four values of $\epsilon$ two values correspond to the non-extreme event regime and two values correspond to the extreme event regime. In both the cases, the performance was adequate for all the DL models. To choose the best model among the three, the RMSE value for each model and for each value of $\epsilon$ were calculated and compared using a bar plot. From the comparison, we found that the LSTM is more accurate in predicting the extreme events than the other two models. Further, the actual values and the predicted values were fitted and it was further confirmed that the fit produced by LSTM is more accurate among the three. LSTM also performs very well in multi-step forecasting of the time series. From the obtained results, we conclude that for the system (\ref{pardri}), LSTM can serve as the best model for the model-free prediction of emergence of extreme events. With such an accurate prediction, we can deploy safety measures in order to prevent the devastating and aftermath effects of extreme events in the considered model.

\begin{acknowledgements}
	JM thanks RUSA 2.0 project for providing a fellowship to carry out this work. SS  thanks  the  Department  of  Science  and  Technology (DST), Government of India, for support through INSPIRE Fellowship (IF170319). The work of MS forms a part of a research project sponsored by Council of Scientific and Industrial Research (CSIR) under the Grant No. 03(1397)/17/EMR-II. MS also acknowledges the Department of Science and Technology (DST) under PURSE Phase-II for providing financial support in procuring high performance desktop which highly assisted this work.  
\end{acknowledgements}

\section*{Authors Contribution Statement:}
All the authors contributed equally to the preparation of this manuscript. 
\section*{APPENDIX}
In this Appendix, we examine the robustness of the three models, MLP, CNN and LSTM considered in our study. In general the performance of these models depends on model size and data size. So, we analyse these factors and determine the exact necessities of each model for better model-free prediction of extreme events in the considered system.

First we check the ability of these models in predicting more than one time step using multi-step forecasting.

\subsection*{\textbf{1. Multi-step forecasting}}
\par In our main analysis, we were forecasting the future values of both non-extreme and extreme regime's time series by considering only one successive step. In order to check the performance of the considered model with more than one step in the time series, we modify the input data accordingly. That is instead of having one value in the output $(Y)$ (refer Table.~\ref{table1}), we take the output in such a way that it gives more than one value during the process of forecasting. 
\par
\begin{figure}[!ht]
	\centering
	\includegraphics[width=0.5\linewidth]{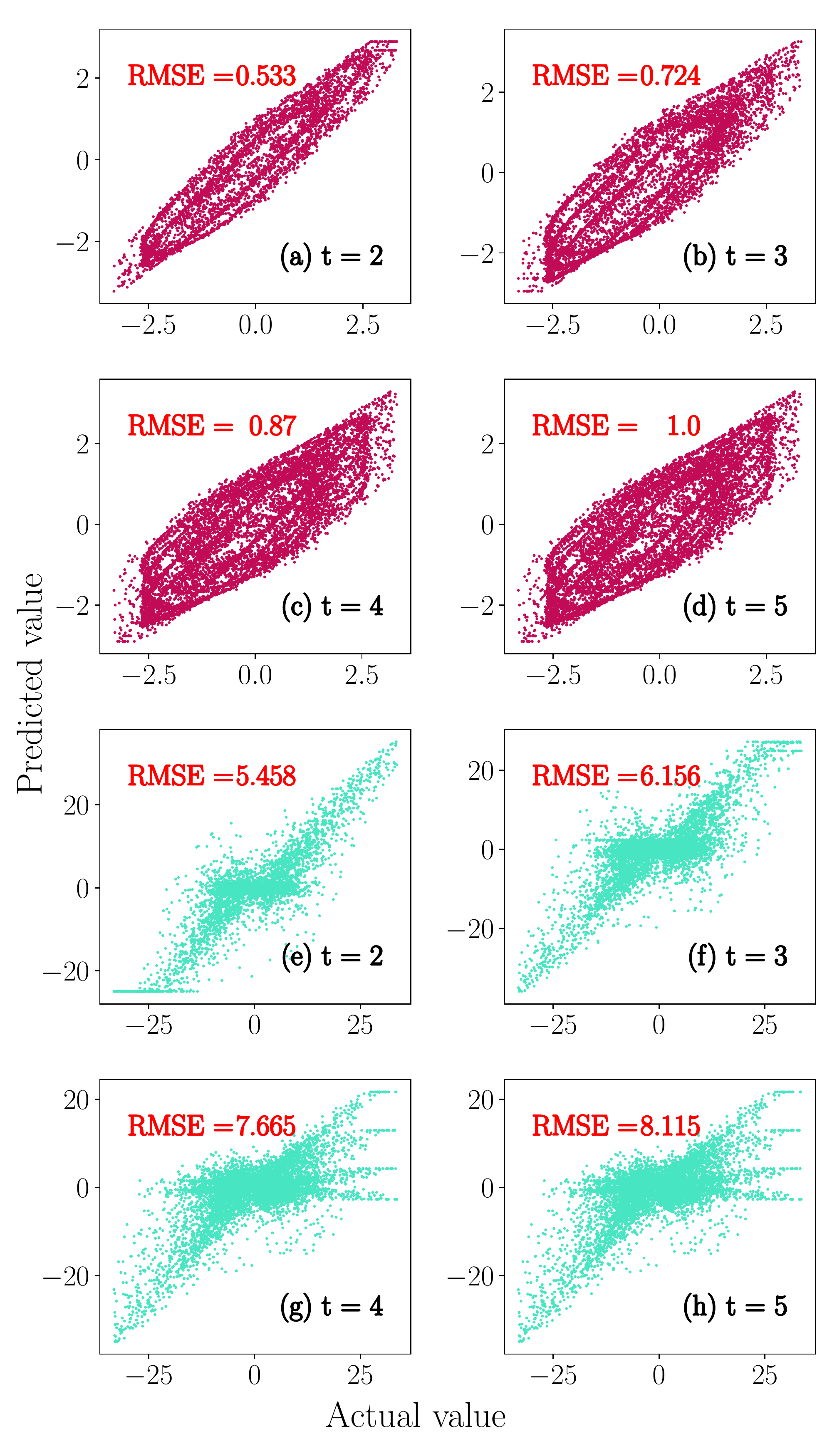}
	\caption{\label{abla-3-mlp} Scattered plots of MLP model for different time steps (t).  (a)-~(d) corresponds to non-extreme case $\epsilon=0.05$ and (e)-~(h) corresponds to the extreme case $\epsilon=0.081$.}
\end{figure}

 The results of MLP model with multi-step forecasting are shown as scattered plots in Fig.~\ref{abla-3-mlp}. The `$t$' in the figure denotes the forecasting time step. From the figure, we can see that in non-extreme regime the RMSE values are in the admissible range but the scattered plots show that the performance of MLP model is low. For extreme case, the RMSE values are very high and the points in the scatter plots are also very much scattered. This shows that MLP's performance in multi-step forecasting of time series with extreme events is low.
\par
\begin{figure}[!ht]
	\centering
	\includegraphics[width=0.5\linewidth]{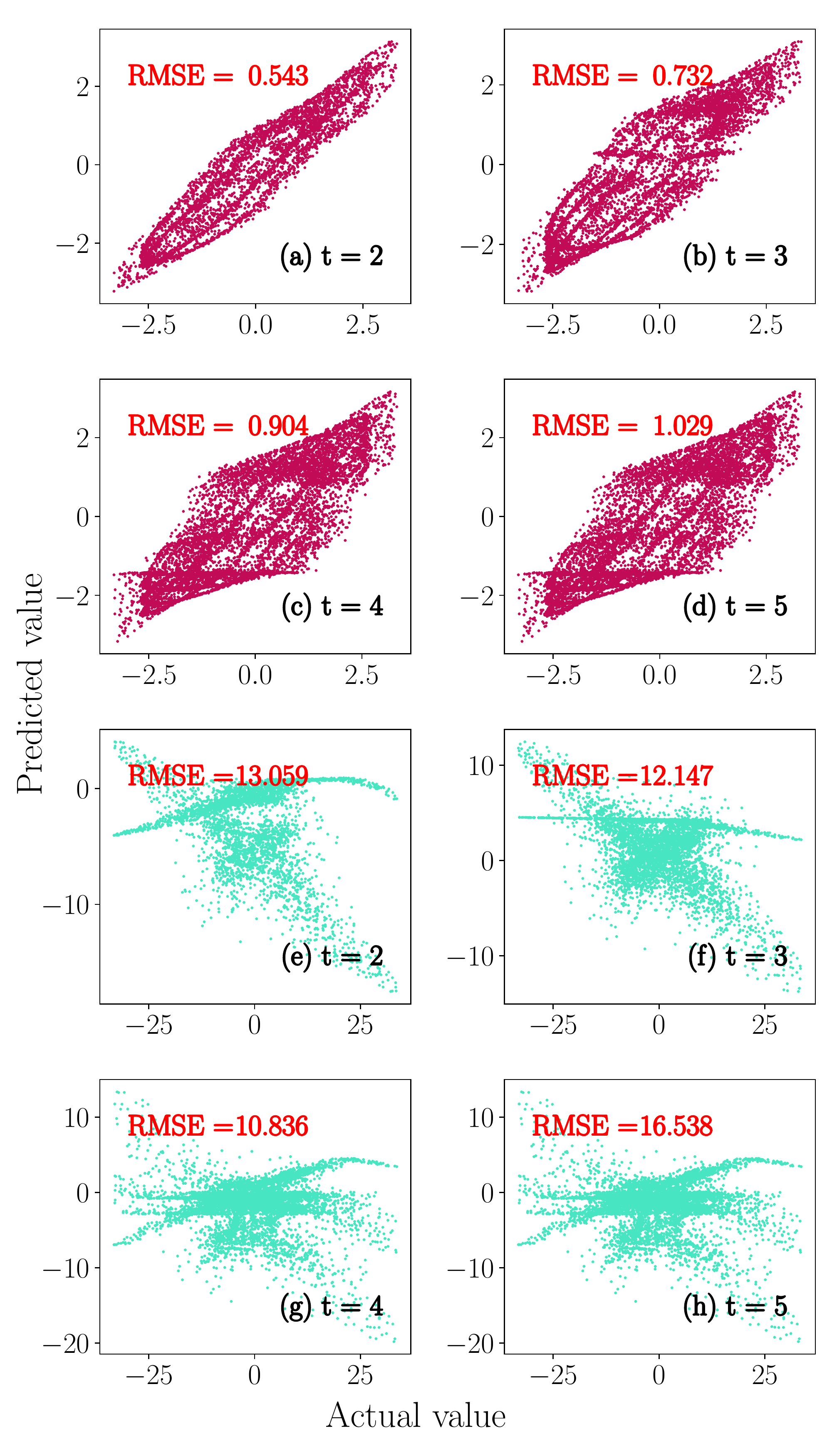}
	\caption{\label{abla-3-cnn} Scattered plots of CNN model for different time steps (t).  ~(a)-~(d) corresponds to non-extreme case $\epsilon=0.05$ and ~(e)-~(h) corresponds to the extreme case $\epsilon=0.081$.}
\end{figure}

 The observations from the experiments of multi-step forecasting using CNN model are shown in Fig.~\ref{abla-3-cnn}. The scatter plots demonstrate the low performance of CNN in multi-step prediction. Moreover, the RMSE values are very high. These two results confirm that for the considered system (1), CNN model also fails in the multi-step forecasting of the time series with extreme events.

\par
\begin{figure}[!ht]
	\centering
	\includegraphics[width=0.5\linewidth]{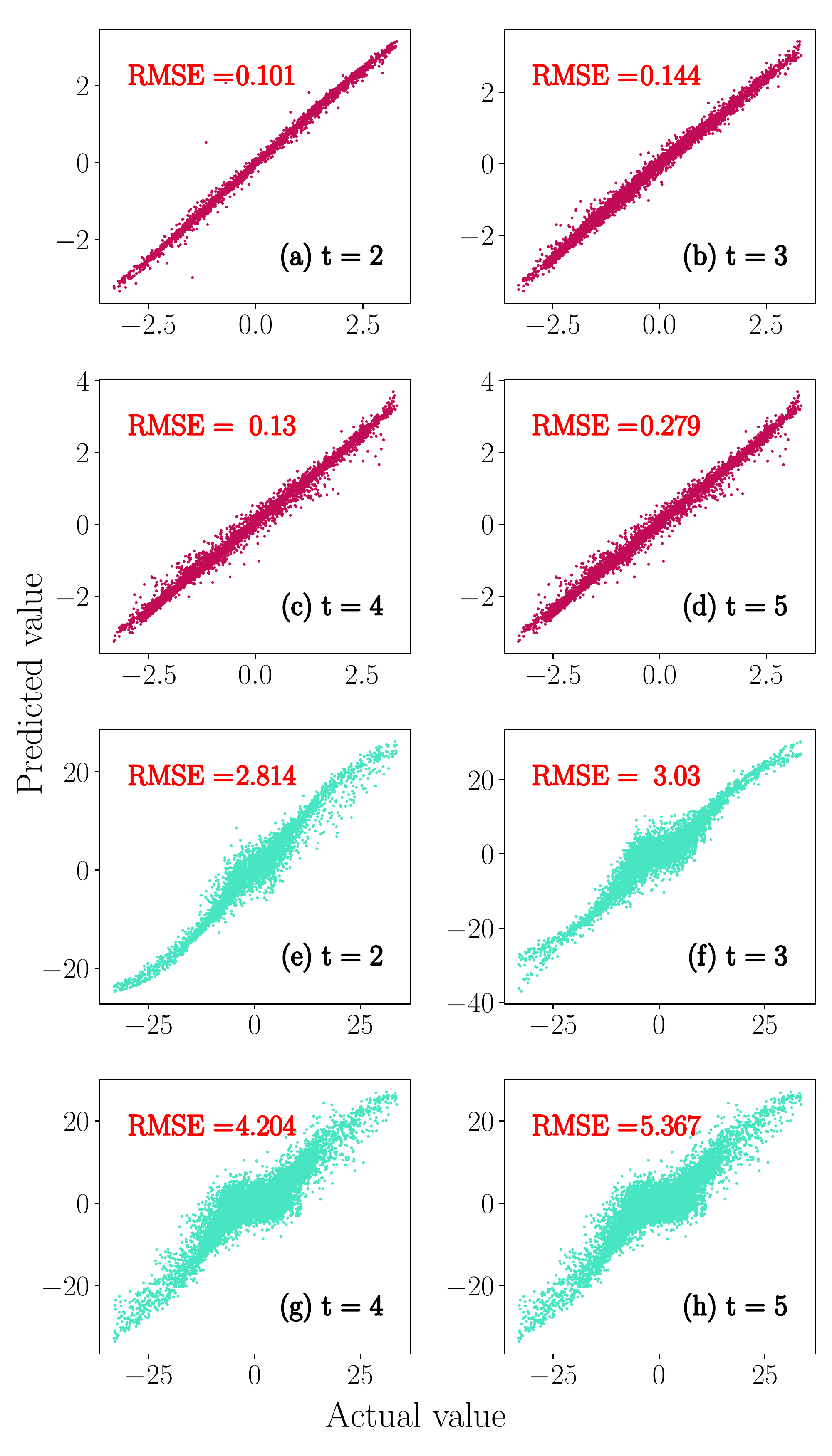}
	\caption{\label{abla-3-lstm} Scattered plots of LSTM model for different time steps (t).  ~(a)-~(d) corresponds to non-extreme case $\epsilon=0.05$ and ~(e)-~(h) corresponds to the extreme case $\epsilon=0.081$.}
\end{figure}

 The results that come out of the multi-step forecasting using LSTM model are shown in Fig.~\ref{abla-3-lstm}. Differing from the above two models, here we can see that the RMSE values for non-extreme case are too good and the scatter plots also confirm that LSTM performs well and good in multi-step forecasting typically up to 5 steps. The forecasting results in extreme regime (Figs.~\ref{abla-3-lstm} (e)-(h)) also ensures that LSTM can do well in multi-step forecasting even upto 5 steps.
\par From the Figs. \ref{abla-3-mlp},\ref{abla-3-cnn} and \ref{abla-3-lstm}, we can infer that LSTM model has a better ability to forecast multiple steps in the time series when compared with the other two models.
\subsection*{\textbf{2. Effect of architecture size}}
Next we analyze the effect of the components of the architecture such as number of layers, number of neurons and kernel size to evaluate the performance of the models.
\subsubsection*{\textbf{2.1 MLP}} 
\par
\begin{figure}[!ht]
	\centering
	\includegraphics[width=0.5\linewidth]{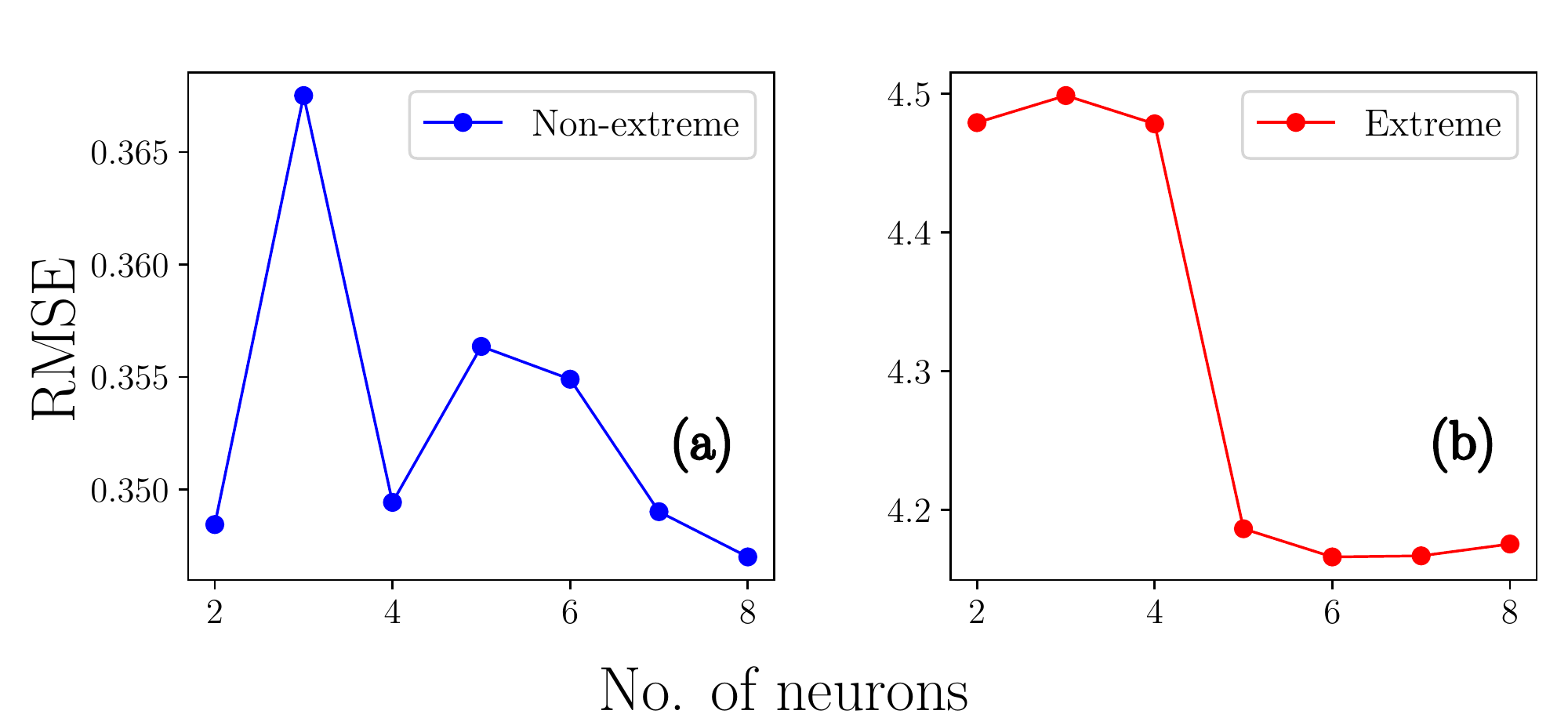}
	\caption{\label{abla-1-mlp} RMSE values for different number of neurons in the 2nd hidden layer of MLP model. (a) corresponds to non-extreme case $\epsilon=0.05$ and (b) corresponds to the extreme case $\epsilon=0.081$.}
\end{figure}

 We checked and found that this model fails to predict the test data with one hidden layer. So we start the prediction of the test set time series with two hidden layers. To see the effect of number of neurons on the performance of the considered MLP model, we fix eight neurons in the first hidden layer and change the number of neurons in the second hidden layer. The variation in the RMSE values with respect to the number of neurons are plotted in Fig.~\ref{abla-1-mlp}. 
In this figure, blue and red dots correspond to the non-extreme ($\epsilon=0.05$) and extreme ($\epsilon=0.081$) cases respectively. From Fig.~\ref{abla-1-mlp}, we notice that the value of RMSE decreases when we increase the number of neurons. In non-extreme regime, the changes in the RMSE values are low (0.365 to 0.35) whereas in the extreme regime there is a significant change in the RMSE values (4.5 to 4.15). This is due to the influence of number of neurons. This confirms that the proposed MLP model performs better with 2 hidden layers each in the size of 8, particularly for forecasting time series that indicates the emergence of extreme events in the considered system (\ref{pardri}).

\subsubsection*{\textbf{2.2 CNN}} 
\par
\begin{figure}[!ht]
	\centering
	\includegraphics[width=0.5\linewidth]{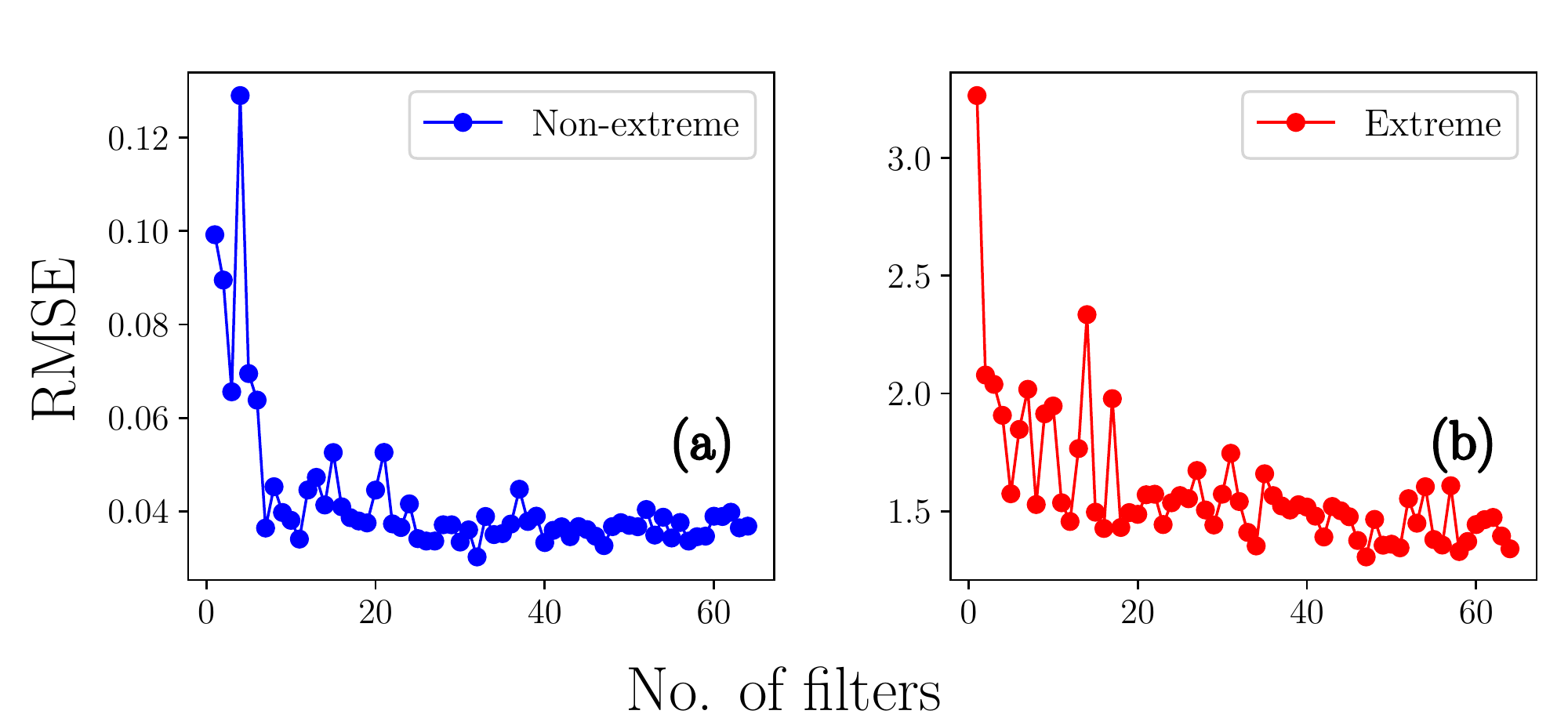}
	\caption{\label{abla-1-cnn} RMSE values for different number of filters in the 1D convolutional layer of CNN model. (a) corresponds to non-extreme case $\epsilon=0.05$ and (b) corresponds to the extreme case $\epsilon=0.081$.}
\end{figure}

 We proposed a CNN model with one 1D convolutional layer. In the following, we study the changes in the performance of the considered model with respect to the number of filters with kernel size of two. For this purpose, we slightly modify the input in such a way that instead of having only one value in the input, now we take 5 values. In Fig.~\ref{abla-1-cnn}, we plot the RMSE values of test set data against the number of filters. 
In the figure blue and red dots respectively corresponds to the non-extreme ($\epsilon=0.05$) and extreme ($\epsilon=0.081$) cases. In both the cases the performance is increased while increasing the number of filters. The variation of RMSE values in the non-extreme case is low whereas the variation of RMSE values in the extreme case is high. From the observed results, we note that the performance of the CNN model in predicting the extreme time series depends on the number of filters in it.

\subsubsection*{\textbf{2.3 LSTM}} 
\par
\begin{figure}[!ht]
	\centering
	\includegraphics[width=0.5\linewidth]{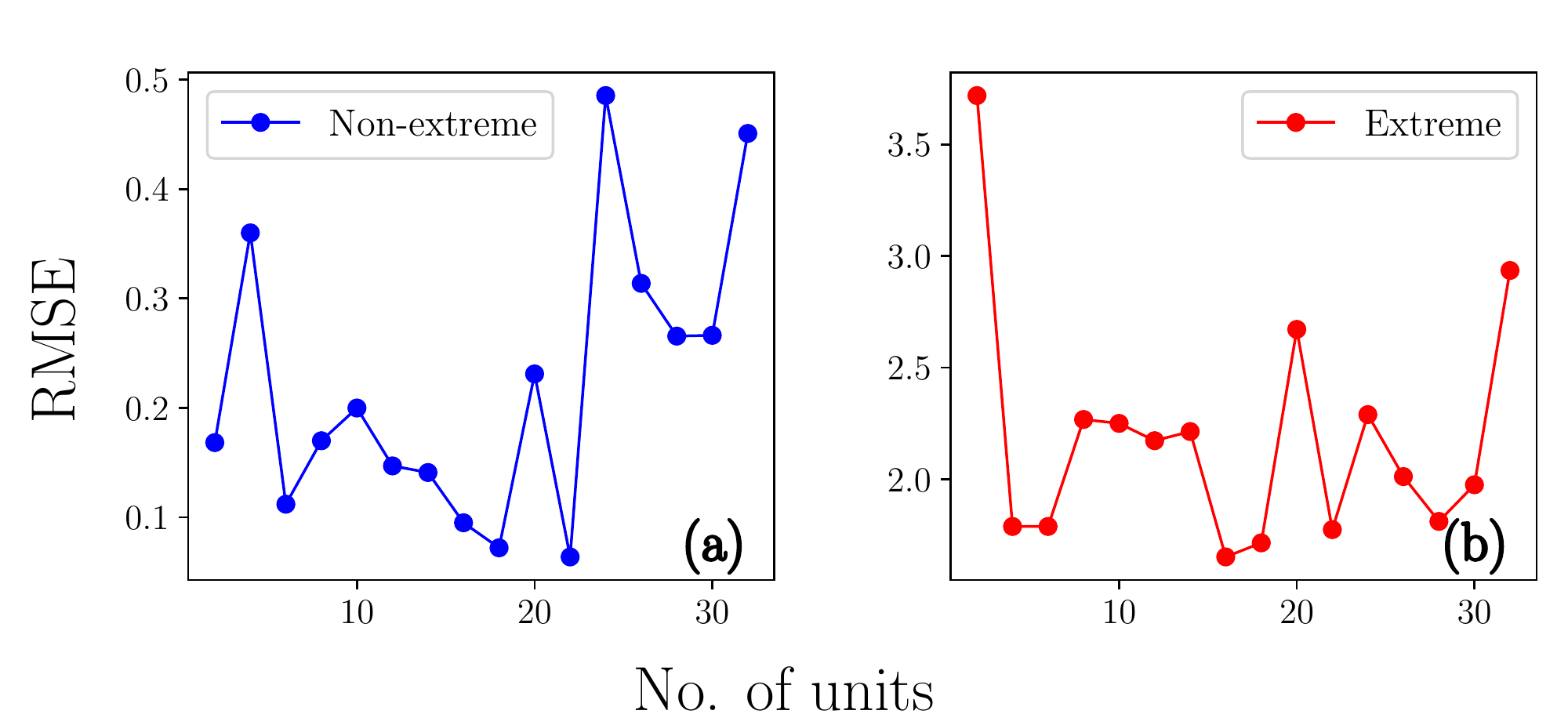}
	\caption{\label{abla-1-lstm1} RMSE values for different number of LSTM units of one LSTM layer. (a) corresponds to non-extreme case $\epsilon=0.05$ and (b) corresponds to the extreme case $\epsilon=0.081$.}
\end{figure}

\begin{figure}[!ht]
	\centering
	\includegraphics[width=0.5\linewidth]{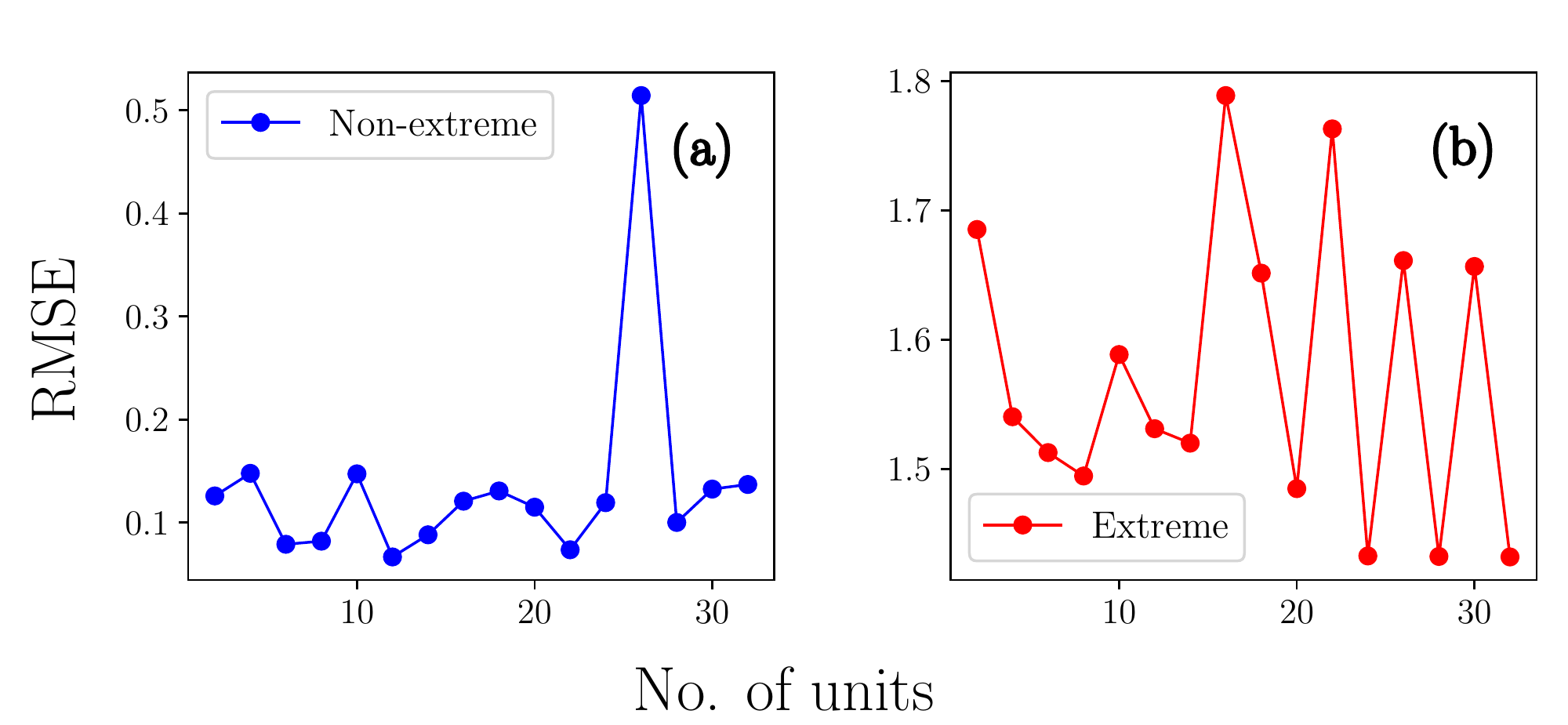}
	\caption{\label{abla-1-lstm2} RMSE values for different number of LSTM units of second LSTM layer. (a) corresponds to non-extreme case $\epsilon=0.05$ and (b) corresponds to the extreme case $\epsilon=0.081$.}
\end{figure}

 In our analysis, we have considered two LSTM layers each with 32 units. To study the effect of model size on the performance of the LSTM we change the number of units in the LSTM layer. 
Now we test the performance of the model with only one LSTM layer. The observed results with one layer is plotted in Fig.~\ref{abla-1-lstm1}. RMSE values go up and down for both the non-extreme (blue dots) and extreme (red dots) cases while changing the number of units. The RMSE value lies between 0.1 and 0.5 for non-extreme case and 1.8 to 3.7 for extreme case.
To know the effect of the second LSTM layer, we fix the first layer with 32 units and change the number of units in the second layer.
The observations are plotted in the Fig.~\ref{abla-1-lstm2}.
From this figure, we can infer that the decrease in RMSE values for extreme case is better when compare to the results obtained with only one layer.

\subsection*{\textbf{3. Effect of data size}}
\par
\begin{figure}[!ht]
	\centering
	\includegraphics[width=0.5\linewidth]{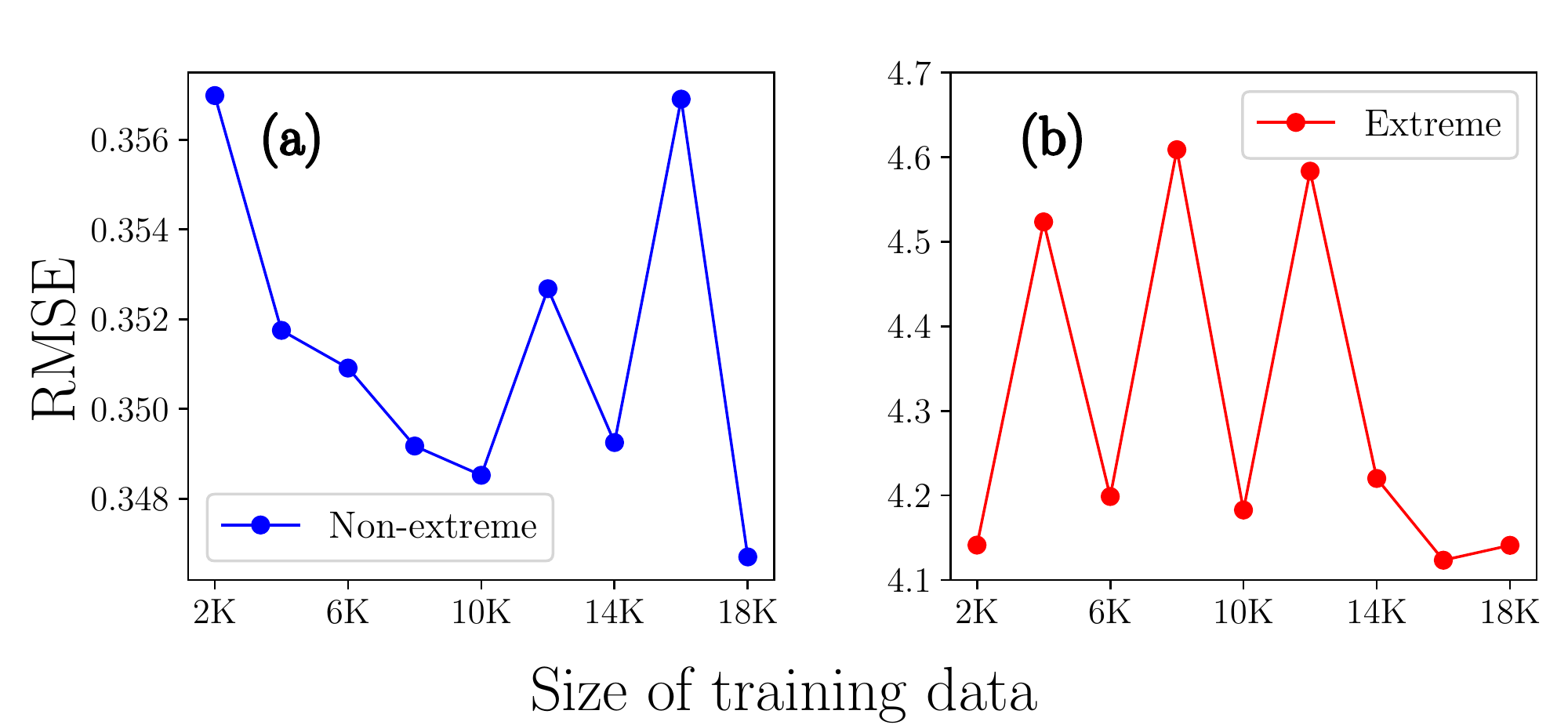}
	\caption{\label{abla-2-mlp} RMSE values for different number of data size of MLP model. (a) corresponds to non-extreme case $\epsilon=0.05$ and (b) corresponds to the extreme case $\epsilon=0.081$.}
\end{figure}

 In order to check the robustness of the model due to the training size data, we train the model with different sizes of the data and testing with 2000 data. We compare the RMSE values obtained from the observations by plotting RMSE against training data size. Figure.~\ref{abla-2-mlp} shows the outcome of the MLP model.
In this figure, we can see that for the forecasting of non-extreme time series, the size of the training data does not affect the performance of the model so much but in the extreme event prediction the RMSE values are not stable until 14000 training data. From these observations we can fix how much data is necessary for the extreme event prediction for the MLP model. In our case, we need atleast 14000 training data.
\begin{figure}[!ht]
	\centering
	\includegraphics[width=0.5\linewidth]{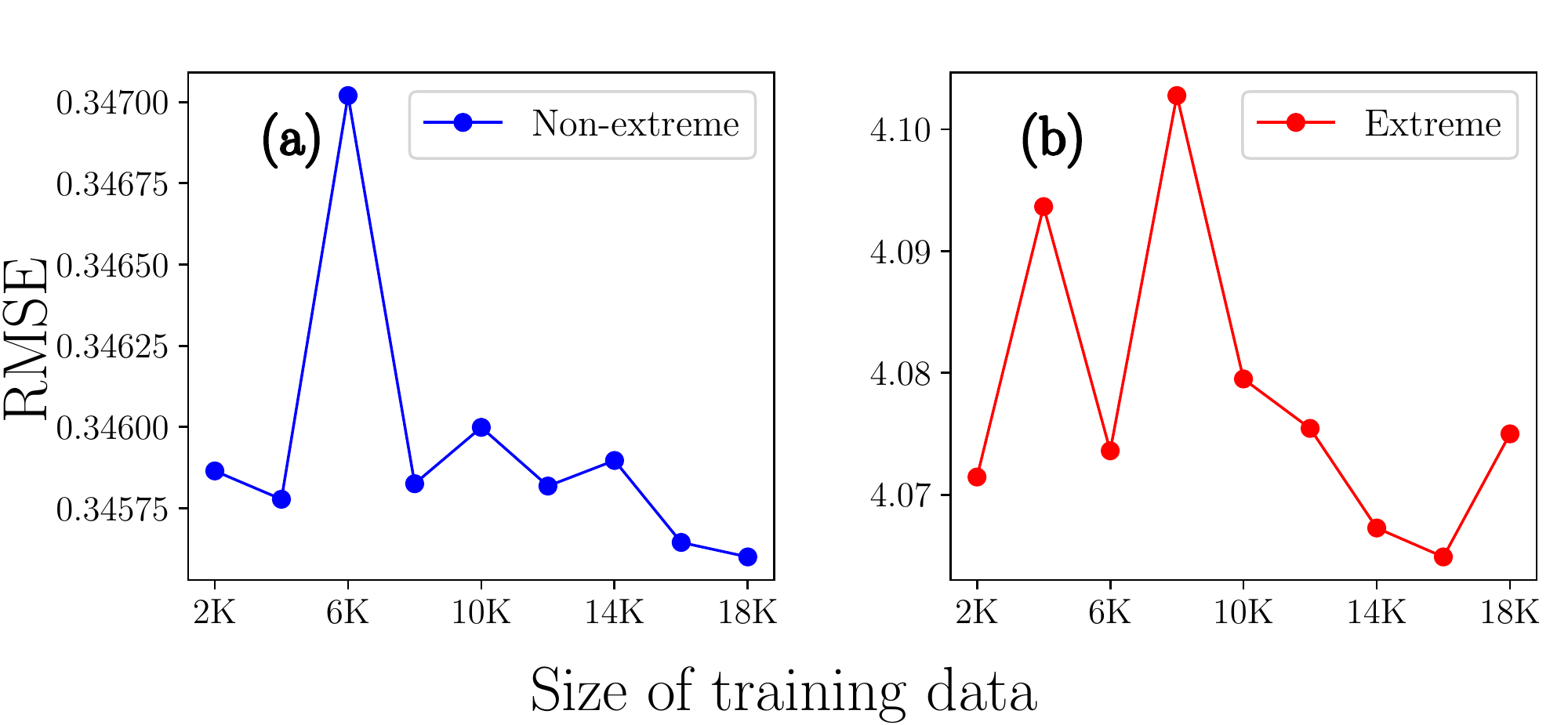}
	\caption{\label{abla-2-cnn} RMSE values for different number of data size of CNN model. (a) corresponds to non-extreme case $\epsilon=0.05$ and (b) corresponds to the extreme case $\epsilon=0.081$.}
\end{figure}
\begin{figure}[!ht]
	\centering
	\includegraphics[width=0.5\linewidth]{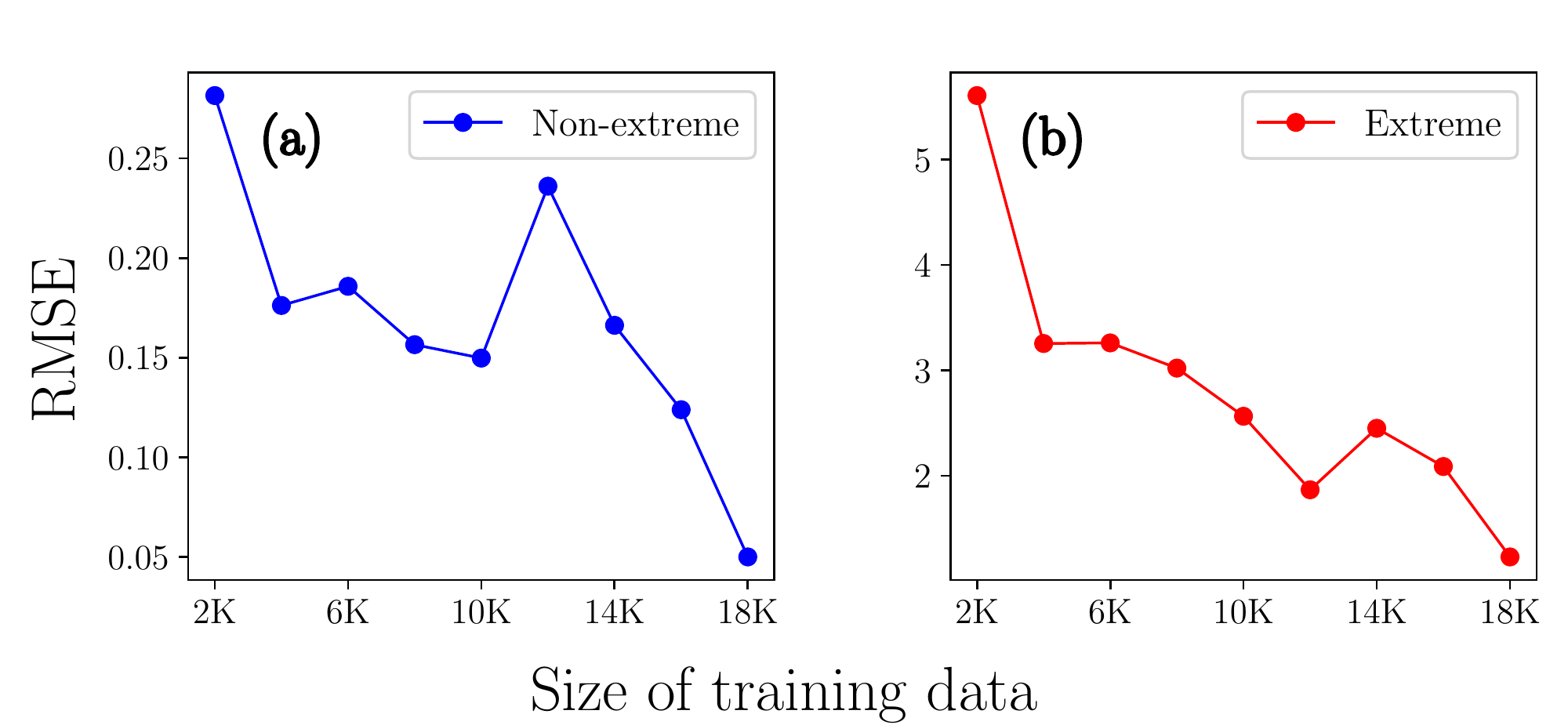}
	\caption{\label{abla-2-lstm} RMSE values for different number of data size of LSTM model. (a) corresponds to non-extreme case $\epsilon=0.05$ and (b) corresponds to the extreme case $\epsilon=0.081$.}
\end{figure}
\par Similarly for CNN, the observations are shown in Fig.~\ref{abla-2-cnn}. From Fig.~\ref{abla-2-cnn} it is clear that in the non-extreme regime the changes in the RMSE values due to the size of the training data is very less (0.345 to 0.347) but in the extreme case the RMSE values go up and down until the data size of 10000. There is only a slight difference in RMSE values (4.6 to 4.10). This behaviour of the CNN confirms that the  performance of CNN for predicting the emergence of extreme event does not depend much on the size of training data.  

\par The performance of predicting the extreme events using LSTM model also depends on the size of the data. The results that come out from the experiment with different sizes of the data are shown in Fig.~\ref{abla-2-lstm}. From Fig.~\ref{abla-2-lstm}, we can see that the RMSE values of non-extreme case start to decrease after the data size crosses 12000 and in the extreme case the RMSE values keep on decreasing after crossing the data size of 2000. Even though it is decreasing from the beginning in extreme regime, the values of the RMSE ranges from 1.3 to 5.6 such that there is a significant decrease in the RMSE value at each step. From the experiment on different sizes of the data, we conclude that to get a better performance with low RMSE value we need to train the LSTM model with sufficient number of data.

\par All the robustness studies performed here will help us to fix the model's architecture appropriately such that the performance of the model in forecasting the extreme events in the considered system is at its best.     

\bibliography{ref.bib}
\end{document}